\documentclass[10pt,twocolumn,letterpaper]{article}

\usepackage[pagenumbers]{cvpr} 

\usepackage{multirow}
\usepackage{siunitx}
\usepackage{setspace} 
\usepackage{comment}
\usepackage{colortbl}  
\usepackage{amsmath}

\DeclareMathOperator*{\argmin}{arg\,min}

\usepackage[dvipsnames]{xcolor}

\definecolor{yzycolor}{rgb}{0.96,0.57,0.58}

\definecolor{sytcolor}{rgb}{0.5,0.67,0.89}

\definecolor{yzybest}{rgb}{0.96, 0.57, 0.58}
\definecolor{yzysecond}{rgb}{0.98, 0.78, 0.57}
\definecolor{yzythird}{rgb}{1.0, 1.0, 0.56}

\definecolor{cvprblue}{rgb}{0.21,0.49,0.74}
\usepackage[pagebackref,breaklinks,colorlinks,citecolor=cvprblue]{hyperref}

\title{SC-GS: Sparse-Controlled Gaussian Splatting for Editable Dynamic Scenes}

\author{
\textbf{Yi-Hua Huang}\textsuperscript{1\#*}
\ 
\textbf{Yang-Tian Sun}\textsuperscript{1\#*}
\ 
\textbf{Ziyi Yang}\textsuperscript{3*}
\ 
\textbf{Xiaoyang Lyu}\textsuperscript{1}
\ 
\textbf{Yan-Pei Cao}\textsuperscript{2$\dagger$}
\ 
\textbf{Xiaojuan Qi}\textsuperscript{1$\dagger$} \\
\textsuperscript{1} The University of Hong Kong
\quad
\textsuperscript{2} VAST
\quad
\textsuperscript{3} Zhejiang University
}

\begin{document}

\twocolumn[{%
\renewcommand\twocolumn[1][]{#1}%
\maketitle

\begin{center}
    \centering
    \captionsetup{type=figure}
    \includegraphics[width=1\linewidth, trim=0 0 0 0, clip]{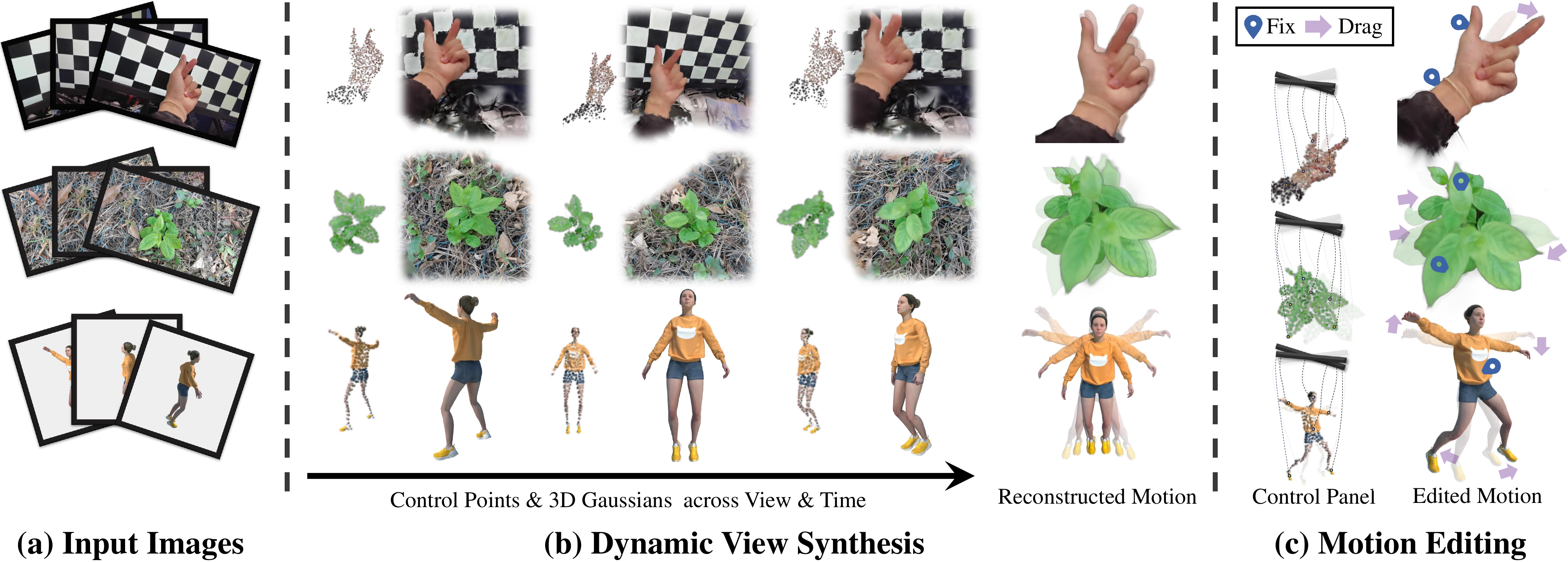}
    \vspace{-6mm}
    \captionof{figure}{Given \textbf{(a)} an image sequence from a monocular dynamic video, we propose to represent the motion with a set of sparse control points, which can be used to drive 3D Gaussians for high-fidelity rendering. Our approach enables both \textbf{(b)} dynamic view synthesis and \textbf{(c)} motion editing due to the motion representation based on sparse control points.
    }
    \vspace{-1mm}
\label{fig:teaser}
\end{center}
}]

\if TT\insert\footins{\noindent\footnotesize{
\#This work is in collaboration with VAST. \\
*Equal Contribution $\dagger$Corresponding Author \\
Project page: \url{https://yihua7.github.io/SC-GS-web/}.}}\fi

\begin{abstract}
\vspace{-4mm}
Novel view synthesis for dynamic scenes is still a challenging problem in computer vision and graphics. Recently, Gaussian splatting has emerged as a robust technique to represent static scenes and enable high-quality and real-time novel view synthesis. Building upon this technique, we propose a new representation that explicitly decomposes the motion and appearance of dynamic scenes into sparse control points and dense Gaussians, respectively. Our key idea is to use sparse control points, significantly fewer in number than the Gaussians, to learn compact 6 DoF transformation bases, which can be locally interpolated through learned interpolation weights to yield the motion field of 3D Gaussians. We employ a deformation MLP to predict time-varying 6 DoF transformations for each control point, which reduces learning complexities, enhances learning abilities, and facilitates obtaining temporal and spatial coherent motion patterns.  Then, we jointly learn the 3D Gaussians, the canonical space locations of control points, and the deformation MLP to reconstruct the appearance, geometry, and dynamics of 3D scenes. During learning, the location and number of control points are adaptively adjusted to accommodate varying motion complexities in different regions, and an ARAP loss following the principle of as rigid as possible is developed to enforce spatial continuity and local rigidity of learned motions. Finally, thanks to the explicit sparse motion representation and its decomposition from appearance, our method can enable user-controlled motion editing while retaining high-fidelity appearances. Extensive experiments demonstrate that our approach outperforms existing approaches on novel view synthesis with a high rendering speed and enables novel appearance-preserved motion editing applications. 
\end{abstract}    
\vspace{-6mm}
\section{Introduction}
\label{sec:intro}

{Novel view synthesis from a monocular video is a crucial problem with many applications in virtual reality, gaming, and the movie industry. However, extracting scene geometry and appearance from limited observations ~\citep{mildenhall2021nerf, wang2021neus, muller2022instant} is challenging. Concurrently, real-world scenes often contain dynamic objects, which pose additional challenges in representing object movements accurately to reflect real-world dynamics~\citep{pumarola2021d, li2021neural, park2021nerfies, park2021hypernerf, li2022neural}.}
Recent advancements in this area are primarily driven by neural radiance fields (NeRF) ~\citep{mildenhall2021nerf,pumarola2021d,li2021neural,yunus2024recent}, which utilizes an implicit function to simultaneously learn scene geometry~\citep{lyu2023learning,liu2023neudf} and textures~\citep{huang2023nerf,yang2023sire} from multi-view images. 
Despite significant progress, NeRF-based representations still struggle with low rendering speeds and high memory usage. This issue is particularly evident when rendering at high resolutions \citep{yu2021plenoctrees,fridovich2022plenoxels,muller2022instant}, as they necessitate sampling hundreds of query points along each ray to predict color and opacity. 

Most recently, Gaussian splatting \cite{kerbl20233d} has shown remarkable performance in terms of rendering quality, resolution, and speed. Utilizing a point-based~\citep{keselman2022fuzzy,xu2022point,dai2020neural, zhang2022differentiable, keselman2023fuzzyplus,gao2023surfelnerf,wang2023neural2} scene representation, this method rasterizes 3D Gaussians to render images from specified views. It enables fast model training and real-time inference, achieving state-of-the-art (SOTA) visual quality. However, its existing formulation only applies to static scenes. It remains a challenge to incorporate object motion into the Gaussian representation without compromising rendering quality and speed. 
An intuitive method is to learn a flow vector for each 3D Gaussian. However, it will incur a significant time cost for training and inference. Moreover, it also leads to noisy trajectories and poor generalization in novel views, as demonstrated in Fig.~\ref{fig:ablation1} (a).

Motivated by the observation that real-world motions are often \textit{sparse}, \textit{spatially continuous}, and \textit{locally rigid}, we propose to drive 3D Gaussians with learnable sparse control points ($\approx$512) compared to the number of Gaussians ($\approx$100K), in a much more compact space for modeling scene dynamics. 
These control points are associated with time-varying 6 DoF transformations parameterized as rotation using quaternion and translation parameters, which can be locally interpolated through learned interpolation weights to yield the motion field of dense Gaussians. 
These 6 DoF parameters on control points are predicted by an introduced MLP conditioned on time and location.
Then, we jointly learn the canonical space 3D Gaussian parameters, locations, and radius of sparse control points at canonical space and the MLP for dynamic novel view synthesis. During learning, we introduce an adaptive strategy to adaptively change the number of sparse points to accommodate motion complexities in different regions and employ an ARAP loss that encourages the learned motions to be locally rigid.

Owing to the effective motion and appearance representations, our approach simultaneously enables high-quality dynamic view synthesis and motion editing, as shown in Fig.~\ref{fig:teaser}. We perform extensive experiments and ablation studies on benchmark datasets, demonstrating that our model surpasses existing methods both quantitatively and qualitatively while maintaining high rendering speeds. 
Furthermore, by learning a control graph from the scene motion, our control points-based motion representation allows for convenient motion editing, a feature not present in existing methods ~\cite{pumarola2021d, TiNeuVox, Cao2023HexPlaneAF, kplanes_2023, guo2023forward}. 
More results for motion editing are included in Fig.~\ref{fig:compare_editing} and the supplementary material.
Our contributions can be summarized as follows:
\begin{itemize}
    \item 
    We introduce sparse control points together with an MLP for modeling scene motion, based on the insight that motions within a scene can be represented by a compact subspace with a sparse set of bases. 
    \item We employ adaptive learning strategies and design a regularization loss based on rigid constraints to enable effective learning of appearances, geometry, and motion from a monocular video.
    \item Thanks to the sparse motion representation, our approach enables motion editing by manipulating the learned control points while maintaining high-fidelity appearances.
    \item Extensive experiments show our approach achieves SOTA performance quantitatively and qualitatively.
\end{itemize}

\section{Related Work}
\label{sec:rela}

\noindent\textbf{Dynamic NeRF.}
Novel view synthesis has been a prominent topic in the academic field for several years. 
NeRF~\citep{mildenhall2021nerf} models static scenes implicitly with MLPs, and many works~\cite{pumarola2021d, li2021neural, xian2021space, tretschk2021non, park2021nerfies, park2021hypernerf, guo2023forward,yunus2024recent} have expanded its usage to dynamic scenes via a deformation field. Some methods~\citep{gao2021dynamic, li2022neural, park2023temporal} represent dynamic scenes as 4D radiance fields but face extensive computational costs due to ray point sampling and volume rendering.
Several acceleration approaches have been used for dynamic scene modeling. DeVRF~\citep{liu2022devrf} introduces a grid representation, and IBR-based methods~\cite{lin2023im4d, li2023dynibar, lin2022efficient, xu20234k4d} use multi-camera information for quality and efficiency. Other methods used primitives~\citep{lombardi2021mixture}, predicted MLP maps~\citep{peng2023representing}, or grid/plane-based structures~\cite{shao2023tensor4d, TiNeuVox, Cao2023HexPlaneAF, kplanes_2023, wang2023mixed, wang2023neural} for speed and performance in various dynamic scenes. However, hybrid models underperform with high-rank dynamic scenes due to their low-rank assumption.

\vspace{0.1in}\noindent\textbf{Dynamic Gaussian Splatting.}
Gaussian Splatting~\citep{kerbl20233d,wu2024recent} offers improved rendering quality and speed for radiance fields. Several concurrent works have adapted 3D Gaussians for dynamic scenes. Luiten \etal~\citep{luiten2023dynamic} utilizes frame-by-frame training, suitable for multi-view scenes. Yang \etal~\citep{yang2023deformable3dgs} separate scenes into 3D Gaussians and a deformation field for monocular scenes but face slow training due to an extra MLP for learning Gaussian offsets. Following \citep{yang2023deformable3dgs}, Wu~\etal~\cite{wu20234dgaussians} replaced the MLP with multi-resolution hex-planes~\cite{Cao2023HexPlaneAF} and a lightweight MLP. Yang \etal~\citep{yang2023gs4d} include time as an additional feature in 4D Gaussians but face quality issues compared to constraints in canonical space. Our work proposes using sparse control points to drive the deformation of 3D Gaussians, which enhances rendering quality and reduces MLP query overhead. The learned control point graph can also be used for motion editing.

\noindent\textbf{3D Deformation and Editing.}
Traditional deformation methods in computer graphics are typically based on Laplacian coordinates~\cite{Lipman2005LaplacianFF, SorkineHornung2004LaplacianSE, SorkineHornung2005LaplacianMP, sorkine2007rigid, gao2019sparse}, Poisson equation~\cite{yu2004mesh} and cage-based approaches~\cite{wang2020NeuralCage, zhang2020proxy}. These methods primarily focus on preserving the geometric details of 3D objects during the deformation process. In recent years, there have been other approaches~\cite{Yuan2022NeRFEditingGE, yuan2023interactive, xu2022deforming,zheng2023editablenerf, lin2023sketchfacenerf} that aim to edit the scene geometry learned from 2D images. These methods prioritize the rendering quality of the edited scene. Our approach falls into this category. However,  instead of relying on the implicit and computationally expensive NeRF-based approach, our method employs an explicit point-based control graph deformation strategy and Gaussian rendering, which is more intuitive and efficient.

\begin{figure*}[htb]
\begin{center}  
\includegraphics[width=0.9\linewidth]{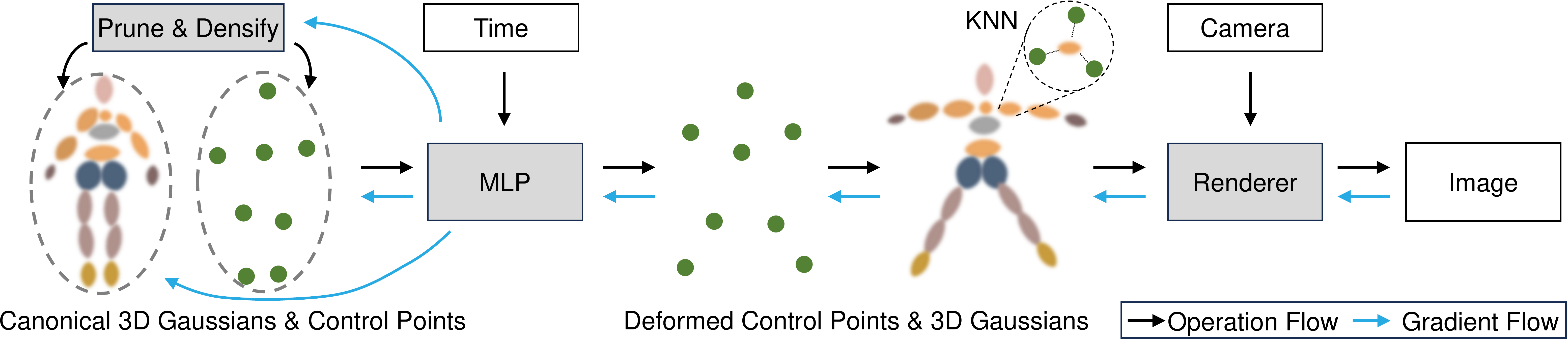}
\vspace{-6mm}
\end{center}
\caption{We present a novel method of employing sparse control points and a deformation MLP to direct 3D Gaussian dynamics. The MLP uses canonical control point coordinates and time to obtain per-control-point 6-DOF transformations, which guide 3D Gaussian deformation based on K nearest control points. Transformed Gaussians can then be rendered into images, and rendering loss calculated, before backpropagating gradients to optimize the Gaussians, control points, and MLP. Gaussian and control point density are adaptively managed during training.}
\label{fig:pipeline}
\vspace{-6mm}
\end{figure*}

\section{Preliminaries}
\label{subsec:method_pre}
Gaussian splatting represents a 3D scene using colored 3D Gaussians~\cite{kerbl20233d}. Each Gaussian $G$ has a 3D center location $\mu$ and a 3D covariance matrix $\Sigma$, 
\begin{equation}
G(x) = e^{-\frac{1}{2} (x-\mu)^T \Sigma^{-1}(x-\mu)}.
\end{equation}
The covariance matrix $\Sigma$ is decomposed as $\Sigma = RSS^TR^T$ for optimization, with $R$ as a rotation matrix represented by a quaternion $q \in \mathbf{SO}(3)$, and $S$ as a scaling matrix represented by a 3D vector $s$. Each Gaussian has an opacity value $\sigma$ to adjust its influence in rendering and is associated with sphere harmonic (SH) coefficients $sh$ for view-dependent appearance. A scene is parameterized as a set of Gaussians $\mathcal{G} = \{G_j: \mu_j, q_j, s_j, \sigma_j, sh_j \}$.

Rendering an image involves projecting these Gaussians onto the 2D image plane and aggregating them using fast $\alpha$-blending. The 2D covariance matrix and center are $\Sigma^{\prime} = JW\Sigma W^{T}J^{T}$ and $\mu^{\prime} = JW\mu$. The color $C(u)$ of a pixel $u$ is rendered using a neural point-based $\alpha$-blending as, 
\begin{equation}
\label{equa:gaussian_render}
\small
C ({u}) = \sum_{i \in N} T_i \alpha_i \mathcal{SH}(sh_i, v_i), \text{ where } T_i = \Pi_{j=1}^{i - 1}(1 - \alpha_{j}).
\end{equation}
Here, $\mathcal{SH}$ is the spherical harmonic function and $v_i$ is the view direction. $\alpha_i$ is calculated by evaluating the corresponding projected Gaussian $G_i$ at pixel $u$ as, \begin{equation}
\label{equa:render2}
\alpha_i = \sigma_i e^{-\frac{1}{2} ({p} - \mu_i^{\prime})^T \Sigma_i^{\prime} ({p} - \mu_i^{\prime}) },
\end{equation}
where $\mu_i^{\prime}$ and $\Sigma_i^{\prime}$ are the center point and covariance matrix of Gaussian $G_i$, respectively.
By optimizing the Gaussian parameters $\{G_j: \mu_j, q_j, s_j, \sigma_j, c_j \}$ and adjusting Gaussian density adaptively, high-quality images can be synthesized in real-time. We further introduce sparse control points to adapt Gaussian splatting for dynamic scenes while maintaining rendering quality and speed.

\section{Method}
\label{sec:method}

Our goal is to reconstruct a dynamic scene from a monocular video.
We represent the geometry and appearance of the dynamic scene using Gaussians in the canonical space
while \emph{modeling the motion through a set of control points together with time-varying 6DoF transformations predicted by an MLP}. These learned control points and corresponding transformations can be utilized to drive the deformation of Gaussians across different timesteps.
The number of control points is significantly smaller than that of Gaussians, resulting in a set of \emph{compact} motion bases for modeling scene dynamics and further facilitating \emph{motion editing}. 
An overview of our method is shown in Fig.~\ref{fig:pipeline}.
In the following, we first present the sparse control points for representing compact motion bases in Sec.~\ref{subsec:method_warp}, followed by the dynamic scene rendering formulation in Sec.~\ref{sec:scenerender} and optimization process in Sec.~\ref{sec:optimization}.

\subsection{Sparse Control Points}
\label{subsec:method_warp}

To derive a compact motion representation, we introduce a set of sparse control points  $\mathcal{P} = \{(p_i \in \mathbb{R}^3, o_i \in \mathbb{R}^+)\}, i\in \{1,2, \cdots, N_p\}$. Here,   $p_i$ denotes the learnable coordinate of control point $i$ in the canonical space. $o_i$ is the learnable radius parameter of a radial-basis-function (RBF) kernel that controls how the impact of a control point on a Gaussian will decrease as their distances increase. $N_p$ is the total number of control points, which is considerably fewer than that of Gaussians.

For each control point $k$, we learn time-varying 6 DoF transformations  $[R_{i}^t|T_{i}^t] \in \mathbf{SE}(3)$ , consisting of a local frame rotation matrix $R_{i}^t \in \mathbf{SO}(3)$ and a translation vector $T_{i}^t \in \mathbb{R}^3$. 
Instead of directly optimizing the transformation parameters for each control point at different time steps, we employ an MLP $\Psi$ to learn a time-varying transformation field and query the transformation of each control point $p_k$ at each timestep $t$ as: 
\begin{equation}
    \label{Equa:control_points} 
    \Psi: (p_i, t) \rightarrow (R_{i}^t, T_{i}^t). 
\end{equation}
Note that in practical implementations, $R_{i}^t$ is represented equivalently as a quaternion $r_{i}^t$ for more stable optimization and convenient interpolation for generating the motions of Gaussians in the follow-up steps. 

\subsection{Dynamic Scene Rendering} 
\label{sec:scenerender}
Equipped with the time-varying transformation parameters $(R_{i}^t, T_{i}^t)$ for sparse control points which form a set of compact motion bases, the next step is to determine the transformation of each Gaussian at different time steps to derive the motion of the entire scene. 
We derive the dense motion field of Gaussians using linear blend skinning (LBS)~\citep{sumner2007embedded} by locally interpolating the transformations of their neighboring control points. Specifically, 
for each Gaussian $G_j:(\mu_j, q_j, s_j,\sigma_j,sh_j)$, we use k-nearest neighbor (KNN) search to obtain its $K(=4)$ neighboring control points denoted as $\{p_{k} | k \in \mathcal{N}_j \}$ in canonical space. Then, the interpolation weights for control point $p_{k}$ can be computed with Gaussian-kernel RBF~\citep{gao2019surfelwarp,dou2016fusion4d,newcombe2015dynamicfusion} as:
\begin{equation}
\small
\label{eq:control_weight}
    w_{jk} = \frac{\hat{w}_{jk}}{\sum\limits_{k \in \mathcal{N}_j} \hat{w}_{jk}} \text{, where } \ \hat{w}_{jk} = \text{exp}(-\frac{d_{jk}^2}{2o_k^2}), 
\end{equation}
where $d_{jk}$ is the distance between center of Gaussian $G_j$ and the neighboring control point $p_k$, and $o_k$ is the learned radius parameter of $p_k$.   During training, these interpolation weights are adaptable to model complex motions by encouraging the learnable radius parameters to be optimized in a way that can accurately reconstruct the video frames.  

Using the interpolation weights of neighboring control points, we can calculate a Gaussian motion field through interpolation. 
Following dynamic fusion works~\citep{newcombe2015dynamicfusion,li2009robust,dou2016fusion4d}, we employ LBS~\citep{sumner2007embedded} to compute the warped Gaussian $\mu_j^t$ and $q_j^t$ as Eq.~\eqref{Equa:warp_gaussian_pos} and Eq.~\eqref{Equa:warp_gaussian_pos2} for simplicity and efficiency:
\begin{align}
    \label{Equa:warp_gaussian_pos}
    \mu_j^t &= \sum\limits_{k \in \mathcal{N}_j} w_{jk} \left(R_{k}^t(\mu_j - p_k) + p_k + T_{k}^t\right), \\
    \label{Equa:warp_gaussian_pos2}
    q_j^t &= (\sum\limits_{k \in \mathcal{N}_j} w_{jk} r_{k}^t) \otimes q_j,
\end{align}
where $R_{k}^t \in \mathbb{R}^{3\times3}$ and $r_k^t \in \mathbb{R}^4$ are the matrix and quaternion representations of predicted rotation on control point $k$ respectively.
$\otimes$ is the production of quaternions, obtaining the quaternion of the composition of corresponding rotation transforms. 
Then, with the updated Gaussian parameters, we are able to perform rendering at time step $t$ following Eq.~\eqref{equa:gaussian_render} and Eq.~\eqref{equa:render2}. 

\subsection{Optimization} 
\label{sec:optimization}
Our dynamic scene representation consists of control points $\mathcal{P}$ and Gaussians $G$ in the canonical space and the deformation MLP $\Psi$. To stabilize the training process, we first pre-train $\mathcal{P}$ and $\Psi$ to model the coarse scene motion with the Gaussians $\mathcal{G}$ fixed. The details are included in the supplementary material. Then, the whole model is optimized jointly. To facilitate learning, we introduce an ARAP loss to encourage the learned motion of control points to be locally rigid and employ an adaptive density adjustment strategy to adapt to varying motion complexities in different areas. 

\vspace{0.1in}\noindent\textbf{ARAP Loss and Overall Optimization Objective.} 
To avoid local minima and regularize the unstructured control points, we introduce an ARAP loss $\mathcal{L}_\text{arap}$ that encourages their motions to be locally rigid, following the principle of being as rigid as possible~\cite{sorkine2007rigid}.
Before computing the ARAP loss for control points, it is necessary to identify the edges that connect them. To avoid linking unrelated points, we opt to connect the points that have closely aligned trajectories in the scene motion.
Specifically, for a control point $p_j$, we firstly calculate its trajectory $p_i^\text{traj}$ that includes its locations across $N_t(=8)$ randomly sampled time steps as: 
\begin{equation}
\label{Eq:trajectory}
    p_i^\text{traj} = \frac{1}{N_t} p_i^{t_1} \oplus p_i^{t_2} \oplus \cdots \oplus p_i^{t_{N_t}},
\end{equation}
where $\oplus$ denotes vector concatenation operation.
Based on the trajectories obtained, we perform ball queries and use all control points $\mathcal{N}_{c_i}$ within a pre-defined radius to define a local area.
Then, to calculate $\mathcal{L}_\text{arap}$, we randomly sample two time steps $t_1$ and $t_2$. For each point $p_k$ within the radius ({\ie} $k\in \mathcal{N}_{c_i}$), its transformed locations with learned translation parameters $T_k^{t_1}$ and $T_k^{t_2}$ are: $p_k^{t_1} = p_k + T_k^{t_1}$ and  $p_k^{t_2} = p_k + T_k^{t_2}$, thus the rotation matrix $\hat{R}_i$ can be estimated following a rigid motion assumption~\cite{sorkine2007rigid} as: 
\begin{equation}
\small
\label{eq:rotation_matrix}
\hat{R}_i = \argmin\limits_{R \in \mathbf{SO}(3)} \sum_{k\in \mathcal{N}_{c_i}}  {w}_{ik} || (p_i^{t_1}-p_k^{t_1}) - {R}(p_i^{t_2}-p_k^{t_2})||^2.
\end{equation}
Here $w_{ik}$ is calculated similarly to $w_{jk}$ in Eq.~\eqref{eq:control_weight} by replacing Gaussian position $\mu_j$ with control point position $p_i$, which weights the contribution of different neighboring points $p_k$ according to their impact on $p_i$. Eq.~\eqref{eq:rotation_matrix} can be easily solved using SVD decomposition according to~\cite{sorkine2007rigid}.  
Then, $\mathcal{L}_\text{arap}$ is designed as,
\begin{equation}
\label{eq:arap_loss}
\footnotesize
\mathcal{L}_\text{arap}(p_i, t_1, t_2) = \sum_{k\in \mathcal{N}_{c_i}}  {w}_{ik} || (p_i^{t_1}-p_k^{t_1}) - \hat{R}_i(p_i^{t_2}-p_k^{t_2})||^2,
\end{equation}   
which evaluates the degree to which the learned motion deviates from the assumption of local rigidity. By penalizing $\mathcal{L}_\text{arap}$, the learned motions are encouraged to be locally rigid.  
The rigid regularization significantly enhances the learned motion with visualizations shown in Fig.~\ref{fig:ablation1}.

 For optimization, besides $L_\text{arap}$, the rendering loss $\mathcal{L}_\text{render}$ is derived by comparing the rendered image at different time steps with ground truth reference images using a combination of $\mathcal L_1$ loss and D-SSIM loss following \cite{kerbl20233d}. Finally, the overall loss is constructed as: $\mathcal{L} = \mathcal{L}_\text{render} + \mathcal{L}_\text{arap}$.

\begin{table*}[th]
  \centering
  \caption{\textbf{Quantitative comparison on D-NeRF~\citep{pumarola2021d} datasets.}
  We present the average PSNR/SSIM/LPIPS (VGG) values for novel view synthesis on dynamic scenes from D-NeRF, with each cell colored to indicate the \colorbox{yzybest}{best}, \colorbox{yzysecond}{second best}, and \colorbox{yzythird}{third best}.
  } 
  \label{tab:recon}
  \vspace{-2mm}
  \setlength\tabcolsep{0pt}
  \resizebox{1.0\textwidth}{!}{
  \footnotesize
    \begin{tabular}{lcccccccccccc }
        \toprule[1pt]
        \multirow{2}{*}{Methods} &
          \multicolumn{3}{c}{\textbf{Hook}} &
          \multicolumn{3}{c}{\textbf{Jumpingjacks}} &
          \multicolumn{3}{c}{\textbf{Trex}} &
          \multicolumn{3}{c}{\textbf{BouncingBalls}} \\
          \cmidrule(lr){2-4}
          \cmidrule(lr){5-7}
          \cmidrule(lr){8-10}
          \cmidrule(lr){11-13}
          & {\footnotesize{PSNR($\uparrow$)}} & {\footnotesize{SSIM($\uparrow$)}} & {\footnotesize{LPIPS($\downarrow$)}} & {\footnotesize{PSNR($\uparrow$)}} & {\footnotesize{SSIM($\uparrow$)}} & {\footnotesize{LPIPS($\downarrow$)}} & {\footnotesize{PSNR($\uparrow$)}} & {\footnotesize{SSIM($\uparrow$)}} & {\footnotesize{LPIPS($\downarrow$)}} & {\footnotesize{PSNR($\uparrow$)}} & {\footnotesize{SSIM($\uparrow$)}} & {\footnotesize{LPIPS($\downarrow$)}} \\
          \midrule
        D-NeRF~\cite{pumarola2021d}
        & {29.25} & {.968} & {.1120}      
        & {32.80} & {.981} & {.0381}
        & {31.75} & {.974} & {.0367}
        & {38.93} & {.987} & {.1074}\\
        TiNeuVox-B~\cite{TiNeuVox}
        & {31.45} & {.971} & {.0569}
        & {\cellcolor{yzythird}34.23} & {.986} & {.0383}
        & {\cellcolor{yzythird}32.70} & {.987} & {.0340}
        & {\cellcolor{yzythird}40.73} & {.991} & {.0472}\\
        Tensor4D~\cite{shao2023tensor4d}
        & {29.03} & {.955} & {.0499}      
        & {24.01} & {.919} & {.0768}
        & {23.51} & {.934} & {.0640}
        & {25.36} & {.961} & {.0411}\\
        K-Planes~\cite{kplanes_2023}
        & {28.59} & {.953} & {.0581}
        & {32.27} & {.971} & {.0389}
        & {31.41} & {.980} & {.0234}
        & {40.61} & {.991} & {.0297} \\
        FF-NVS~\cite{guo2023forward}
        & {\cellcolor{yzythird}32.29} & {.980} & {.0400}      
        & {33.55} & {.980} & {.0300}
        & {30.71} & {.960} & {.0400}
        & {40.02} & {.990} & {.0400}\\
        4D-GS~\cite{wu20234dgaussians}
        & {30.99} & \cellcolor{yzysecond}.990 & {\cellcolor{yzythird}.0248}
        & {33.59} & {\cellcolor{yzythird}.990} & {\cellcolor{yzythird}.0242}
        & {32.16} & {\cellcolor{yzythird}.988} & {\cellcolor{yzythird}.0216}
        & {38.59} & {\cellcolor{yzythird}.993} & {\cellcolor{yzythird}.0267}\\ \hline 
        Baseline
        & \cellcolor{yzysecond}34.47 & \cellcolor{yzysecond}.990 & \cellcolor{yzysecond}.0195
        & \cellcolor{yzysecond}35.74 & \cellcolor{yzysecond}.992 & \cellcolor{yzysecond}.0178 
        & \cellcolor{yzysecond}36.37 & \cellcolor{yzysecond}.994 & \cellcolor{yzysecond}.0103 
        & \cellcolor{yzysecond}41.45 & \cellcolor{yzysecond}.996 & \cellcolor{yzysecond}.0190 \\
        Ours
        & \cellcolor{yzybest}39.87 & \cellcolor{yzybest}.997 & \cellcolor{yzybest}.0076
        & \cellcolor{yzybest}41.13 & \cellcolor{yzybest}.998 & \cellcolor{yzybest}.0067
        & \cellcolor{yzybest}41.24 & \cellcolor{yzybest}.998 & \cellcolor{yzybest}.0046
        & \cellcolor{yzybest}44.91 & \cellcolor{yzybest}.998 & \cellcolor{yzybest}.0166
      \end{tabular}}

    \resizebox{1.0\textwidth}{!}{
    \footnotesize
   \begin{tabular}{lcccccccccccc }
        \toprule[1pt]
        \multirow{2}{*}{Methods} &
          \multicolumn{3}{c}{\textbf{Hellwarrior}} &
          \multicolumn{3}{c}{\textbf{Mutant}} &
          \multicolumn{3}{c}{\textbf{Standup}} &
          \multicolumn{3}{c}{\textbf{\textit{Average}}} \\
          \cmidrule(lr){2-4}
          \cmidrule(lr){5-7}
          \cmidrule(lr){8-10}
          \cmidrule(lr){11-13}
          & {\footnotesize{PSNR($\uparrow$)}} & {\footnotesize{SSIM($\uparrow$)}} & {\footnotesize{LPIPS($\downarrow$)}} & {\footnotesize{PSNR($\uparrow$)}} & {\footnotesize{SSIM($\uparrow$)}} & {\footnotesize{LPIPS($\downarrow$)}} & {\footnotesize{PSNR($\uparrow$)}} & {\footnotesize{SSIM($\uparrow$)}} & {\footnotesize{LPIPS($\downarrow$)}} & {\footnotesize{PSNR($\uparrow$)}} & {\footnotesize{SSIM($\uparrow$)}} & {\footnotesize{LPIPS($\downarrow$)}} \\
          \midrule

        D-NeRF~\cite{pumarola2021d}
        & {25.02} & {.955} & {.0633}
        & {31.29} & {.978} & {.0212}
        & {32.79} & {.983} & {.0241}
        & {31.69} & {.975} & {.0575} \\
        TiNeuVox-B~\cite{TiNeuVox}
        & {28.17} & {\cellcolor{yzythird}.978} & {.0706}
        & {33.61} & {.982} & {.0388}
        & {35.43} & {.991} & {.0230}
        & {33.76} & {.983} & {.0441} \\
        Tensor4D~\cite{shao2023tensor4d}    
        & {\cellcolor{yzythird}31.40} & {.925} & {.0675}
        & {29.99} & {.951} & {.0422}
        & {30.86} & {.964} & {.0214}
        & {27.62} & {.947} & {.0471} \\
        K-Planes~\cite{kplanes_2023}
        & {25.27} & {.948} & {.0775}
        & {33.79} & {.982} & {.0207}
        & {34.31} & {.984} & {.0194}
        & {32.32} & {.973} & {.0382} \\
        FF-NVS~\cite{guo2023forward}
        & {27.71} & {.970} & {.0500}      
        & {34.97} & {.980} & {.0300}
        & {\cellcolor{yzythird}36.91} & {.990} & {.0200}
        & {33.73} & {.979} & {.0357} \\
        4D-GS~\cite{wu20234dgaussians}
        & {31.39} & {.974} & {\cellcolor{yzythird}.0436}      
        & {\cellcolor{yzythird}35.98} & {\cellcolor{yzythird}.996} & {\cellcolor{yzythird}.0120}
        & {35.37} & {\cellcolor{yzythird}.994} & {\cellcolor{yzythird}.0136}
        & {\cellcolor{yzythird}34.01} & {\cellcolor{yzythird}.987} & {\cellcolor{yzythird}.0316}\\ \hline
        Baseline
        & \cellcolor{yzysecond}39.07 & \cellcolor{yzysecond}.982 & \cellcolor{yzysecond}.0350 
        & \cellcolor{yzysecond}41.45 & \cellcolor{yzysecond}.998 & \cellcolor{yzysecond}.0045 
        & \cellcolor{yzysecond}41.04 & \cellcolor{yzysecond}.996 & \cellcolor{yzysecond}.0071 
        & \cellcolor{yzysecond}38.51 & \cellcolor{yzysecond}.992 & \cellcolor{yzysecond}.0162 \\
        Ours
        & \cellcolor{yzybest}42.93 & \cellcolor{yzybest}.994 &\cellcolor{yzybest}.0155
        & \cellcolor{yzybest}45.19 & \cellcolor{yzybest}.999 & \cellcolor{yzybest}.0028
        & \cellcolor{yzybest}47.89 & \cellcolor{yzybest}.999 & \cellcolor{yzybest}.0023
        & \cellcolor{yzybest}43.31 & \cellcolor{yzybest}.997 & \cellcolor{yzybest}.0063 \\
        \bottomrule[1pt]

      \end{tabular}}
  \vspace{-3mm}
\end{table*}

\vspace{0.1in}\noindent\textbf{Adaptive Control Points.}
Following~\citep{kerbl20233d}, we also develop an adaptive density adjustment strategy to add and prune control points, which adjusts their distributions for modeling varying motion complexities, {\eg} areas that exhibit complex motion patterns typically require control points of high densities. 
\textbf{1)} To determine whether a control point $p_i$ should be pruned, we calculate its overall impact $W_i = \sum_{j \in \tilde{\mathcal{N}}_i} w_{ji}$ on the set of Gaussians $j \in \tilde{\mathcal{N}}_i$ whose K nearest neighbors include $p_i$. Then, we prune $p_i$ if $W_i$ is close to zero, indicating little contribution to the motion of 3D Gaussians. 
\textbf{2)} To determine whether a control point $p_k$ should be cloned, we calculate the summation of Gaussian gradient norm with respect to Gaussians in set $\tilde{\mathcal{N}}_k$ as: 
\begin{equation}
\small
g_i=\sum\limits_{j \in \tilde{\mathcal{N}}_i} \tilde{w}_{j} ||\frac{d\mathcal{L}}{d\mu_j}||^2_2 \text{, where } \tilde{w}_{j} = \frac{w_{ji}}{\sum\limits_{j \in \tilde{\mathcal{N}}_k} w_{ji}}.
\end{equation}
A large $g_k$ indicates poor reconstructions. Therefore, we clone $p_k$ and add a new control point $p'_k$ to the expected position of related Gaussians to improve the reconstruction:
\begin{equation}
p'_k=\sum\limits_{j \in \tilde{\mathcal{N}}_k} \tilde{w}_{i} \mu_j; \ \sigma'_k = \sigma_k.
\end{equation}

\section{Motion Editing}
\label{subsec:method_animation}
Since our approach utilizes an explicit and sparse motion representation, it further allows for efficient and intuitive motion editing through the manipulation of control points.
It is achieved by predicting the trajectory of each control point across different time steps, determining their neighborhoods, constructing a rigid control graph, and performing motion editing by graph deformation.

\vspace{0.1in}\noindent\textbf{Control Point Graph.} 
With the trained control points $\mathcal{P}$ and the MLP $\Psi$, we construct a control point graph $\mathcal{G}$ that connects control points based on their trajectories. 
For each vertex of the graph, i.e., control point $p_i$, we firstly calculate its trajectory $p_i^\text{traj}$ derived from Eq.~\eqref{Eq:trajectory}.
Then, the vertex is connected with other vertices that fall within a ball of a pre-determined radius parameter based on $p_i^\text{traj}$. 
The edge weights $w_{ij}$ between two connected vertices $p_i$ and $p_j$ are calculated using Eq.~\eqref{eq:control_weight}.
Building the control graph based on point trajectory helps take into account the overall motion sequence instead of a single timestep, which avoids unreasonable edge connections.
We demonstrate the advantage of this choice in the supplementary material.

\vspace{0.1in}\noindent\textbf{Motion Editing.}
In order to maintain the local rigidity, we perform ARAP~\cite{sorkine2007rigid} deformation on the control graph based on constraints specified by users.
Mathematically, given a set of user-defined handle points $\{h_{l} \in \mathbb{R}^3\ | l \in \mathcal{H} \subset \{1, 2, \cdots, N_p\} \}$, 
the control graph $\mathcal{P}^{\prime}$ can be deformed by minimizing the APAR energy formulated as:
\vspace{-3mm}
\begin{equation}
\small
E(\mathcal{P}^{\prime}) =  \sum\limits_{i=1}^{N_p}\sum\limits_{j \in {\mathcal{N}}_i} {w}_{ij} || (p_i^{\prime}-p_j^{\prime}) - \hat{R}_i(p_i-p_j)||^2,
\end{equation}
with the fixed position condition $p_l^{\prime}=h_l$ for $l \in \mathcal{H}$. Here $\hat{R}_i$ is the rigid local rotation defined on each control point.
This optimization problem can be efficiently solved by alternately optimizing local rotations $\hat{R}_i$ and deformed control point positions $p^\prime$. We refer the readers to ~\cite{sorkine2007rigid} for the specific optimization process. 
The solved rotation $\hat{R}_i$ and translation $\hat{T}_i = p_i^\prime - p_i$ form a 6 DoF transformation for each control point, which is consistent with our motion representation. Thus, Gaussians can be warped by the deformed control points by simply replacing the transformation in Eq.~\eqref{Equa:warp_gaussian_pos} and Eq.~\eqref{Equa:warp_gaussian_pos2}, which can be rendered into high-quality edited images even for motion out of the training sequence.
We visualize the motion editing results in Fig.~\ref{fig:compare_editing}.

\begin{table*}[ht]
  \centering
  \caption{\textbf{Quantitative comparison on NeRF-DS~\citep{yan2023nerf} datasets.} 
  We display the average PSNR/MS-SSIM/LPIPS (Alex) metrics for novel view synthesis on dynamic scenes from NeRF-DS, with each cell colored to indicate the \colorbox{yzybest}{best}, \colorbox{yzysecond}{second best}, and \colorbox{yzythird}{third best}.
  } \label{tab:nerf_ds_table}
  \vspace{-3mm}
  \setlength\tabcolsep{0pt}
  \resizebox{1.0\textwidth}{!}{
  \footnotesize
    \begin{tabular}{lcccccccccccc}
        \toprule[1pt]
        \multirow{2}{*}{Methods} &
          \multicolumn{3}{c}{\textbf{Bell}} &
          \multicolumn{3}{c}{\textbf{Sheet}} &
          \multicolumn{3}{c}{\textbf{Press}} &
          \multicolumn{3}{c}{\textbf{Basin}} \\
          \cmidrule(lr){2-4}
          \cmidrule(lr){5-7}
          \cmidrule(lr){8-10}
          \cmidrule(lr){11-13}
          & {\footnotesize{PSNR($\uparrow$)}} & {\footnotesize{MS-SSIM($\uparrow$)}} & {\footnotesize{LPIPS($\downarrow$)}} & {\footnotesize{PSNR($\uparrow$)}} & {\footnotesize{MS-SSIM($\uparrow$)}} & {\footnotesize{LPIPS($\downarrow$)}} & {\footnotesize{PSNR($\uparrow$)}} & {\footnotesize{MS-SSIM($\uparrow$)}} & {\footnotesize{LPIPS($\downarrow$)}} & {\footnotesize{PSNR($\uparrow$)}} & {\footnotesize{MS-SSIM($\uparrow$)}} & {\footnotesize{LPIPS($\downarrow$)}} \\
          \midrule
        HyperNeRF~\citep{park2021hypernerf}
        & \cellcolor{yzythird}24.0 & \cellcolor{yzythird}.884 & {.159}
        & {24.3} & {.874} & {.148}
        & \cellcolor{yzythird}25.4 & {.873} & {.164}
        & \cellcolor{yzythird}20.2 & {.829} & {.168}         \\
        NeRF-DS~\citep{yan2023nerf}    
        & {23.3} & {.872} & {.134} 
        & \cellcolor{yzythird}25.7 & \cellcolor{yzybest}.918 & \cellcolor{yzybest}.115 
        & \cellcolor{yzysecond}26.4 & \cellcolor{yzybest}.911 & \cellcolor{yzybest}.123
        & \cellcolor{yzysecond}20.3 & \cellcolor{yzysecond}.868 & \cellcolor{yzysecond}.127       \\
        TiNeuVox-B~\citep{TiNeuVox}
        & {23.1} & {.876} & \cellcolor{yzybest}.113
        & {21.1} & {.745} & {.234}
        & {24.1} & \cellcolor{yzythird}.892 & \cellcolor{yzysecond}.133
        & \cellcolor{yzybest}20.7 & \cellcolor{yzybest}.896 & \cellcolor{yzybest}.105         \\ \hline
        Baseline
        & \cellcolor{yzysecond}24.9 & \cellcolor{yzysecond}.917 & \cellcolor{yzythird}.124 
        & \cellcolor{yzysecond}26.1 & \cellcolor{yzysecond}.903 & \cellcolor{yzysecond}.127 
        & {25.1} & {.884} & {.221}
        & {19.6} & \cellcolor{yzythird}.852 & \cellcolor{yzythird}.144          \\
        Ours
        & \cellcolor{yzybest}25.1 & \cellcolor{yzybest}.918 & \cellcolor{yzysecond}.117
        & \cellcolor{yzybest}26.2 & \cellcolor{yzythird}.898 & \cellcolor{yzythird}.142
        & \cellcolor{yzybest}26.6 & \cellcolor{yzysecond}.901 & \cellcolor{yzythird}.135
        & {19.6\textbf{}} & {.846} & {.154}          \\
        \bottomrule[1pt]
        
        \multirow{2}{*}{Methods} &
          \multicolumn{3}{c}{\textbf{Cup}} &
          \multicolumn{3}{c}{\textbf{Sieve}} &
          \multicolumn{3}{c}{\textbf{Plate}} &
          \multicolumn{3}{c}{\textbf{\textit{Average}}} \\
          \cmidrule(lr){2-4}
          \cmidrule(lr){5-7}
          \cmidrule(lr){8-10}
          \cmidrule(lr){11-13}
          & {\footnotesize{PSNR($\uparrow$)}} & {\footnotesize{MS-SSIM($\uparrow$)}} & {\footnotesize{LPIPS($\downarrow$)}} & {\footnotesize{PSNR($\uparrow$)}} & {\footnotesize{MS-SSIM($\uparrow$)}} & {\footnotesize{LPIPS($\downarrow$)}} & {\footnotesize{PSNR($\uparrow$)}} & {\footnotesize{MS-SSIM($\uparrow$)}} & {\footnotesize{LPIPS($\downarrow$)}} & {\footnotesize{PSNR($\uparrow$)}} & {\footnotesize{MS-SSIM($\uparrow$)}} & {\footnotesize{LPIPS($\downarrow$)}} \\
          \midrule
        HyperNeRF~\citep{park2021hypernerf}
        & {20.5} & {.705} & {.318}
        & {25.0} & {.909} & {.129}
        & {18.1} & {.714} & {.359}
        & {22.5} & {.827} & {.206}         \\
        NeRF-DS~\citep{yan2023nerf}    
        & \cellcolor{yzysecond}24.5 & \cellcolor{yzysecond}.916 & \cellcolor{yzythird}.118 
        & \cellcolor{yzybest}26.1 & \cellcolor{yzybest}.935 & \cellcolor{yzybest}.108 
        & \cellcolor{yzybest}20.8 & \cellcolor{yzybest}.867 & \cellcolor{yzysecond}.164
        & \cellcolor{yzysecond}23.9 & \cellcolor{yzybest}.898 & \cellcolor{yzybest}.127          \\
        TiNeuVox-B~\citep{TiNeuVox}
        & {20.5} & {.806} & {.182}
        & {20.1} & {.822} & {.205}
        & \cellcolor{yzysecond}20.6 & \cellcolor{yzysecond}.863 & \cellcolor{yzybest}.161
        & {21.5} & {.843} & {.162}         \\ \hline
        Baseline
        & \cellcolor{yzybest}24.7 & \cellcolor{yzybest}.919 & \cellcolor{yzysecond}.116 
        & \cellcolor{yzythird}25.3 & \cellcolor{yzythird}.917 & \cellcolor{yzysecond}.109 
        & \cellcolor{yzythird}20.3 & \cellcolor{yzythird}.842 & {.214}
        & \cellcolor{yzythird}23.7 & \cellcolor{yzysecond}.891 & \cellcolor{yzythird}.151          \\
        Ours
        & \cellcolor{yzysecond}24.5 & \cellcolor{yzysecond}.916 & \cellcolor{yzybest}.115
        & \cellcolor{yzysecond}26.0 & \cellcolor{yzysecond}.919 & \cellcolor{yzythird}.114
        & {20.2} & {.837} & \cellcolor{yzythird}.202
        & \cellcolor{yzybest}24.1 & \cellcolor{yzysecond}.891 & \cellcolor{yzysecond}.140          \\
        \bottomrule[1pt]
        
      \end{tabular}}
  \vspace{-4mm}
\end{table*}

\section{Experiment}
\label{sec:experiment}

\subsection{Datasets and Evaluation Metrics}
To validate the superiority of our method, we conducted extensive experiments on D-NeRF~\cite{pumarola2021d} datasets and NeRF-DS~\citep{yan2023nerf} datasets.
D-NeRF datasets contain eight dynamic scenes with $360^{\circ}$ viewpoint settings, and the NeRF-DS datasets consist of seven captured videos with camera pose estimated using colmap~\cite{schoenberger2016sfm}. The two datasets involve a variety of rigid and non-rigid deformation of various objects. 
The metrics we use to evaluate the performance are Peak Signal-to-Noise Ratio (PSNR), Structural Similarity(SSIM), Multiscale SSIM(MS-SSIM), and Learned Perceptual Image Patch Similarity (LPIPS)~\cite{Zhang2018TheUE}.

\begin{figure*}[htb]
    \centering
\includegraphics[width=0.9\linewidth]{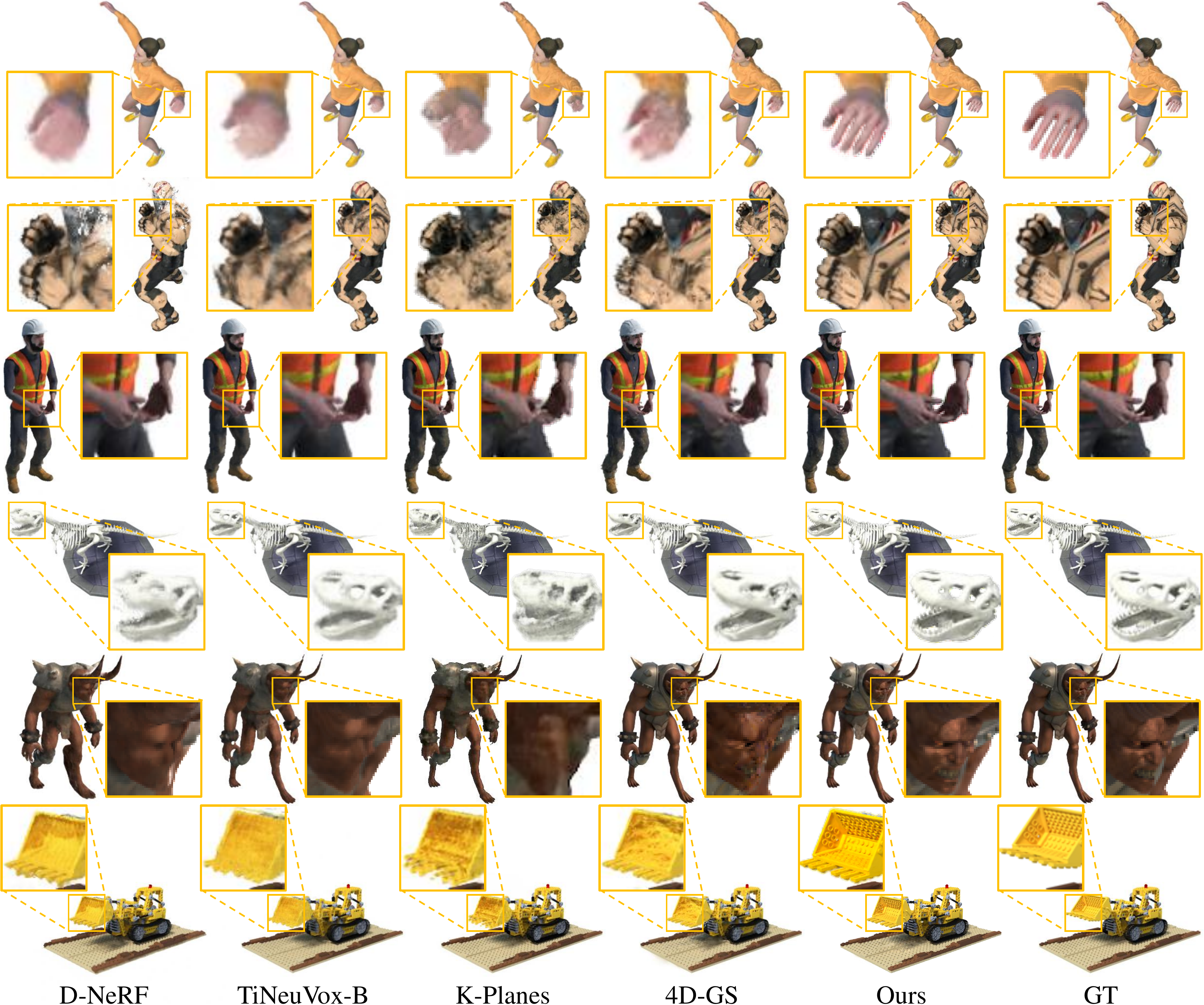}
    \vspace{-1mm}
    \caption{Qualitative comparison of dynamic view synthesis on D-NeRF~\citep{pumarola2021d} datasets. We compare our method with state-of-the-art methods including D-NeRF~\citep{pumarola2021d}, TiNeuVox-B~\citep{TiNeuVox}, K-Planes~\citep{kplanes_2023}, and 4D-GS~\citep{yang2023gs4d}. Our method delivers a higher visual quality and preserves more details of dynamic scenes. Notably, in the Lego scene (bottom row), the train motion is inconsistent with the test motion.}
    \vspace{-5mm}
    \label{fig:compare_sota_synthetic}
\end{figure*}

\vspace{-1mm}
\begin{figure}[htbp]
	\centering
	\includegraphics[width=1\linewidth]{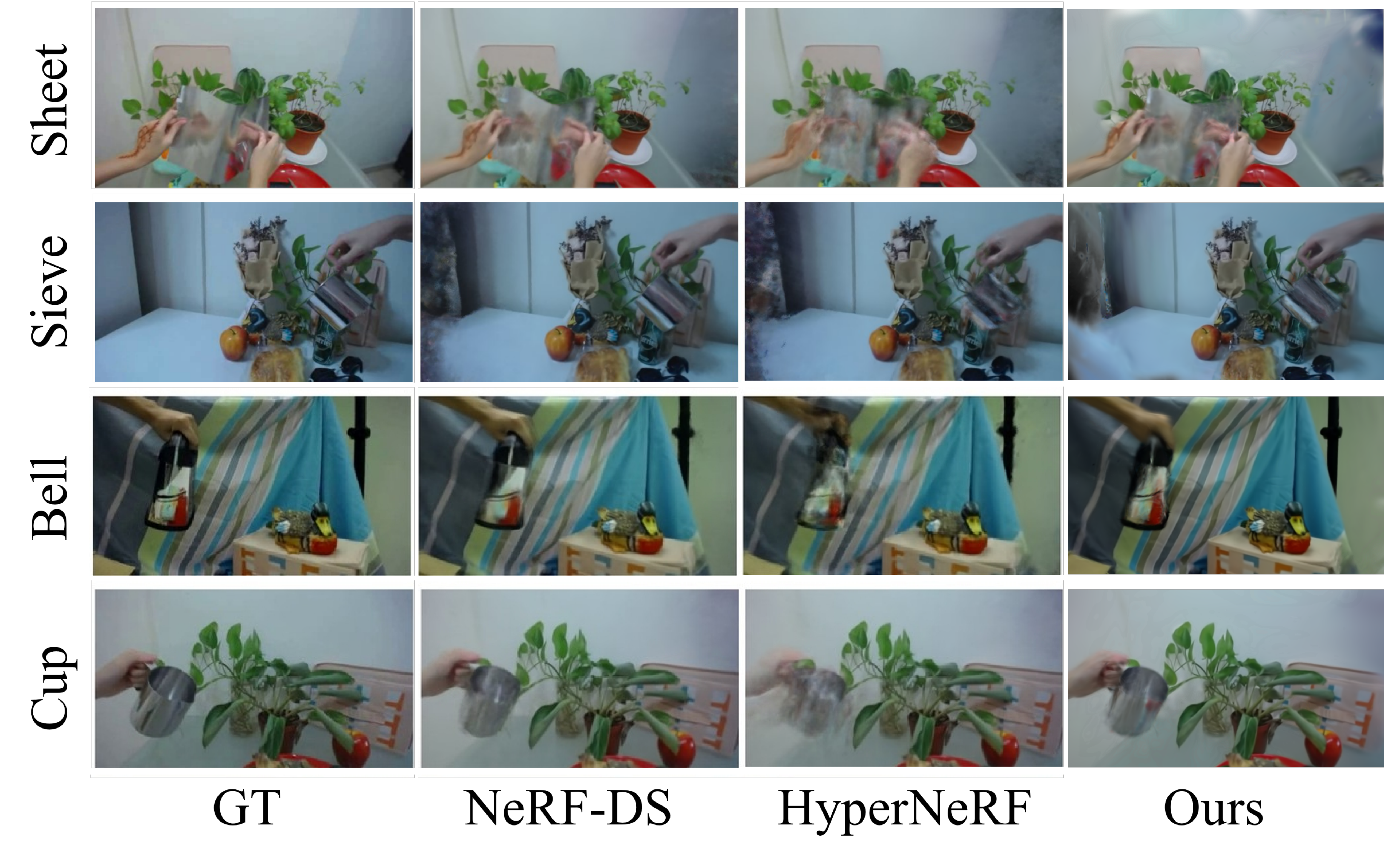}
	\vspace{-6mm}
	\caption{Qualitative comparisons of dynamic view synthesis on scenes from NeRF-DS~\citep{yan2023nerf}. Our method produces high-fidelity results even without specialized design for specular surfaces.}
     \vspace{-4mm}
\label{fig:nerf-ds}
\end{figure}

\subsection{Quantitative Comparisons}
\label{sec:quantitative_comparison}

\noindent\textbf{D-NeRF Datasets.}
We compare our method against existing state-of-the-art methods: D-NeRF~\cite{pumarola2021d}, TiNeuVox~\cite{TiNeuVox}, Tensor4D~\cite{shao2023tensor4d}, K-Planes~\cite{kplanes_2023}, and FF-NVS~\cite{guo2023forward} using the official implementations and follow the same data setting. Concurrent work 4D-GS~\cite{wu20234dgaussians} is also compared since the official code has been released. We also evaluate the baseline that directly applies estimated per-Gaussian transformation with a deformation MLP to demonstrate the effectiveness of control points. The comparisons are carried out on the resolution of 400x400, following the same approach as in previous methods~\cite{pumarola2021d, TiNeuVox, Cao2023HexPlaneAF}. 
We demonstrate the comparison results in Tab.~\ref{tab:recon}. Our approach significantly outperforms others. The baseline method also achieves high synthesis quality thanks to the superiority of 3D Gaussians. However, without the regularization of compact motion bases, the baseline has difficulty in achieving global optima.
We also report the rendering speed comparison in the supplementary material to show the efficiency of our method. 

\noindent\textbf{NeRF-DS Datasets.} Although the datasets provide relatively accurate camera poses compared with \citep{park2021hypernerf}, some inevitable estimation errors still exist. This resulted in a downgraded performance of our method. However, our approach still achieves the best visual quality compared with SOTA methods, as reported in Tab.~\ref{tab:nerf_ds_table}. It's worth mentioning that NeRF-DS outperforms both our method and the baseline on certain datasets, as it employs a specialized design for modeling the specular parts of dynamic objects. Despite this, our approach, which doesn't employ any additional processes, still achieves a higher average performance.


\begin{figure*}[t]
    \centering
    \setlength{\fboxrule}{0.5pt}
    \setlength{\fboxsep}{-0.01cm}
    \setlength\tabcolsep{0pt}
    \begin{spacing}{1}
    \begin{tabular}{p{0.04\linewidth}<{\centering}p{0.16\linewidth}<{\centering}p{0.16\linewidth}<{\centering}p{0.16\linewidth}<{\centering}p{0.16\linewidth}<{\centering}p{0.16\linewidth}<{\centering}p{0.16\linewidth}<{\centering}}

    \rotatebox{90}{ \hspace{-3mm} \small{ Reconstructed Motion}} &
    \includegraphics[width=1\linewidth, trim=50 50 50 50,clip]{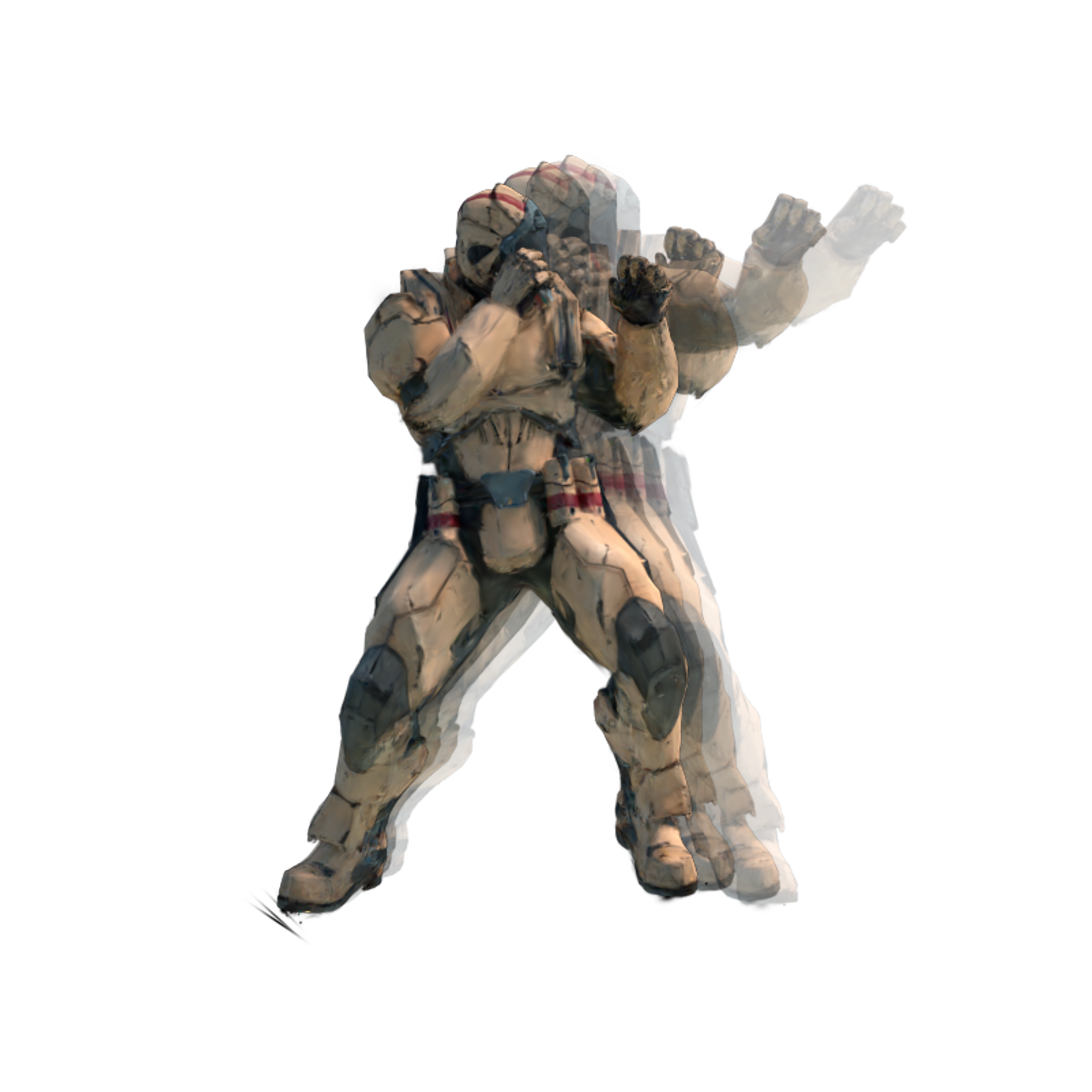} &
    \includegraphics[width=1\linewidth, trim=10 10 10 10,clip]{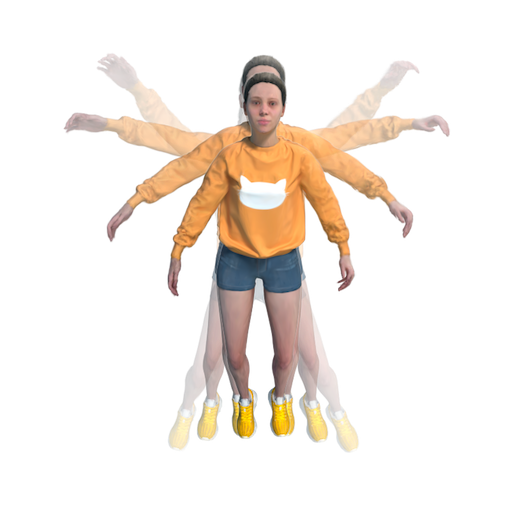} &
    \includegraphics[width=1\linewidth, trim=10 50 50 10,clip]{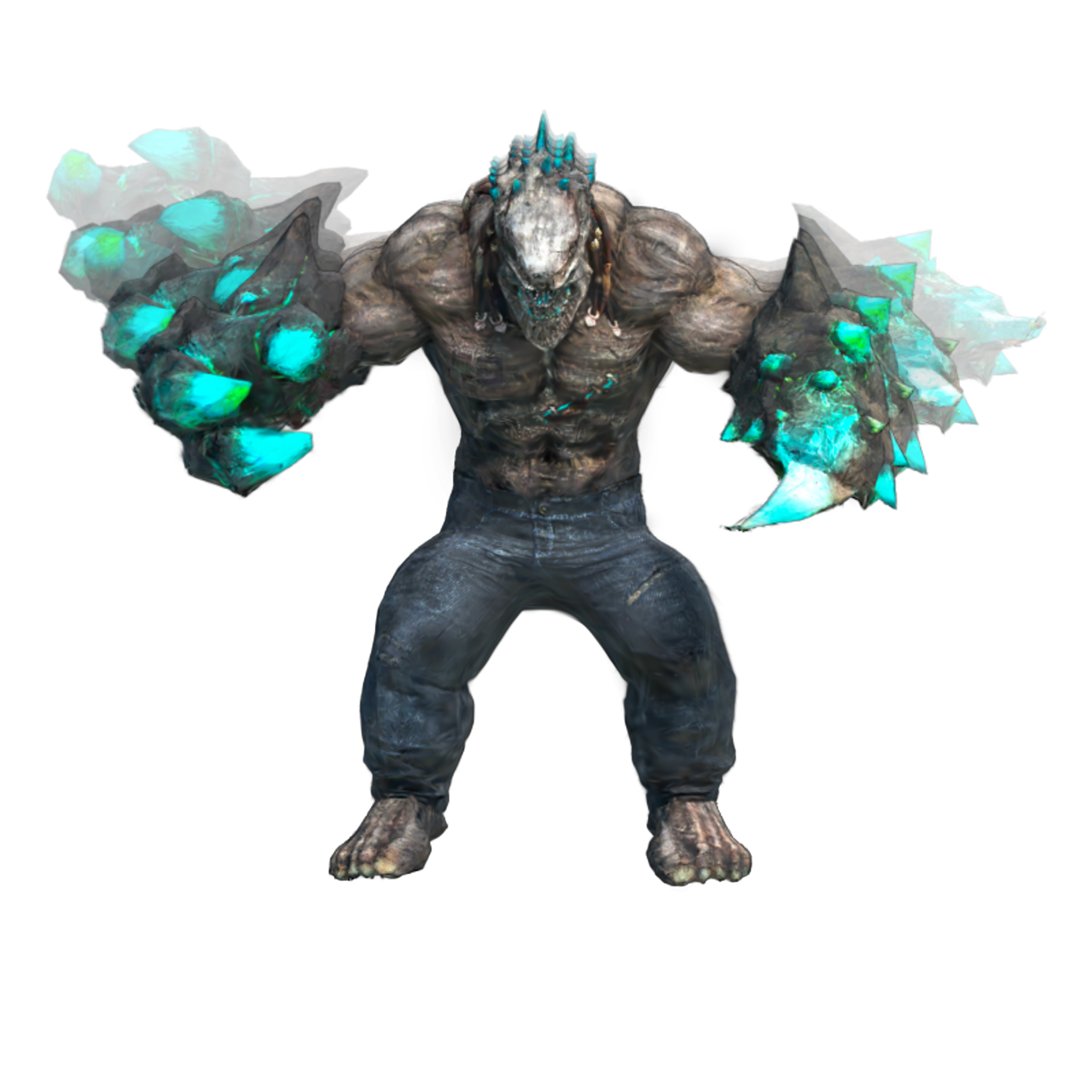} &
    \includegraphics[width=1\linewidth, trim=10 10 10 10,clip]{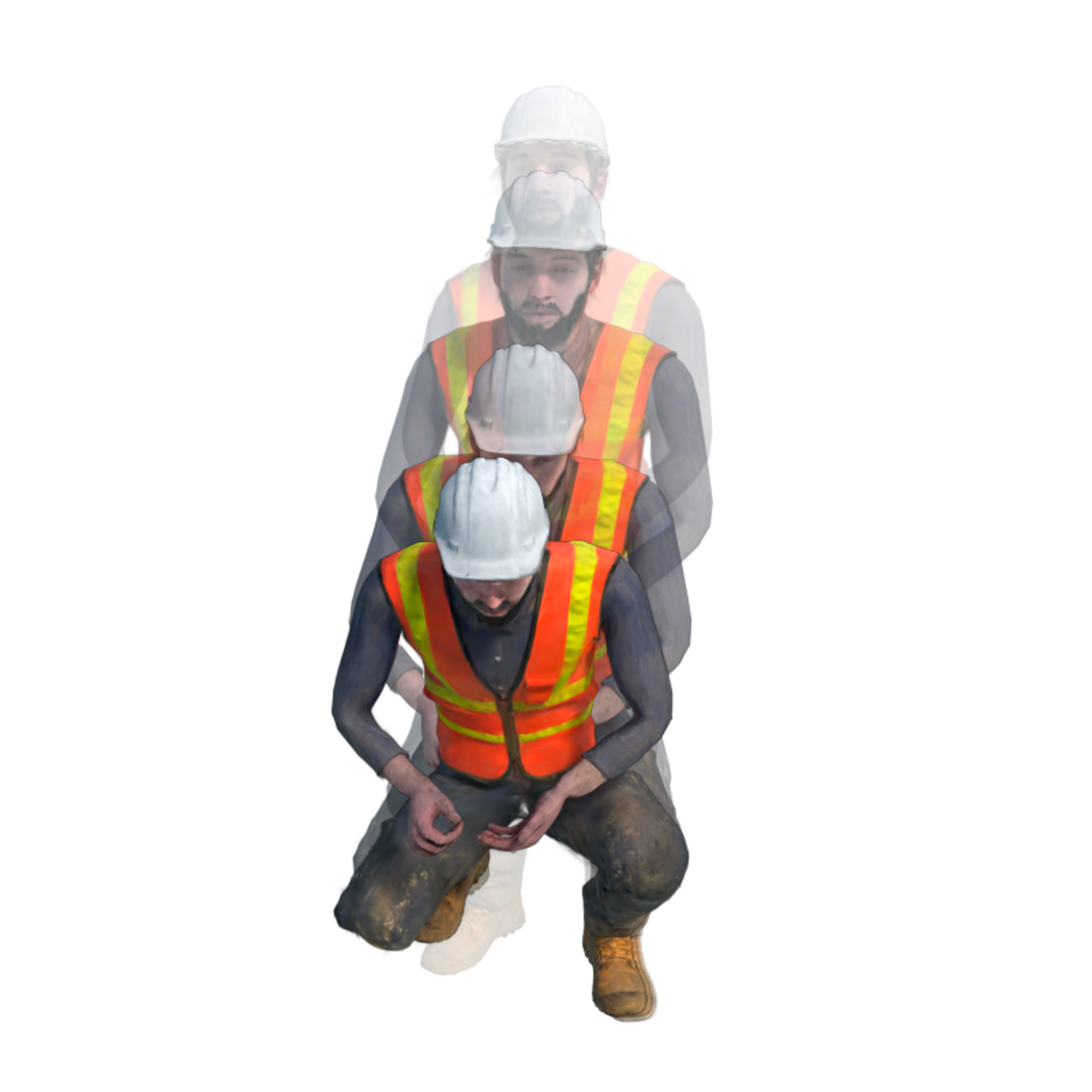} &
    \includegraphics[width=1\linewidth, trim=60 60 10 10,clip]{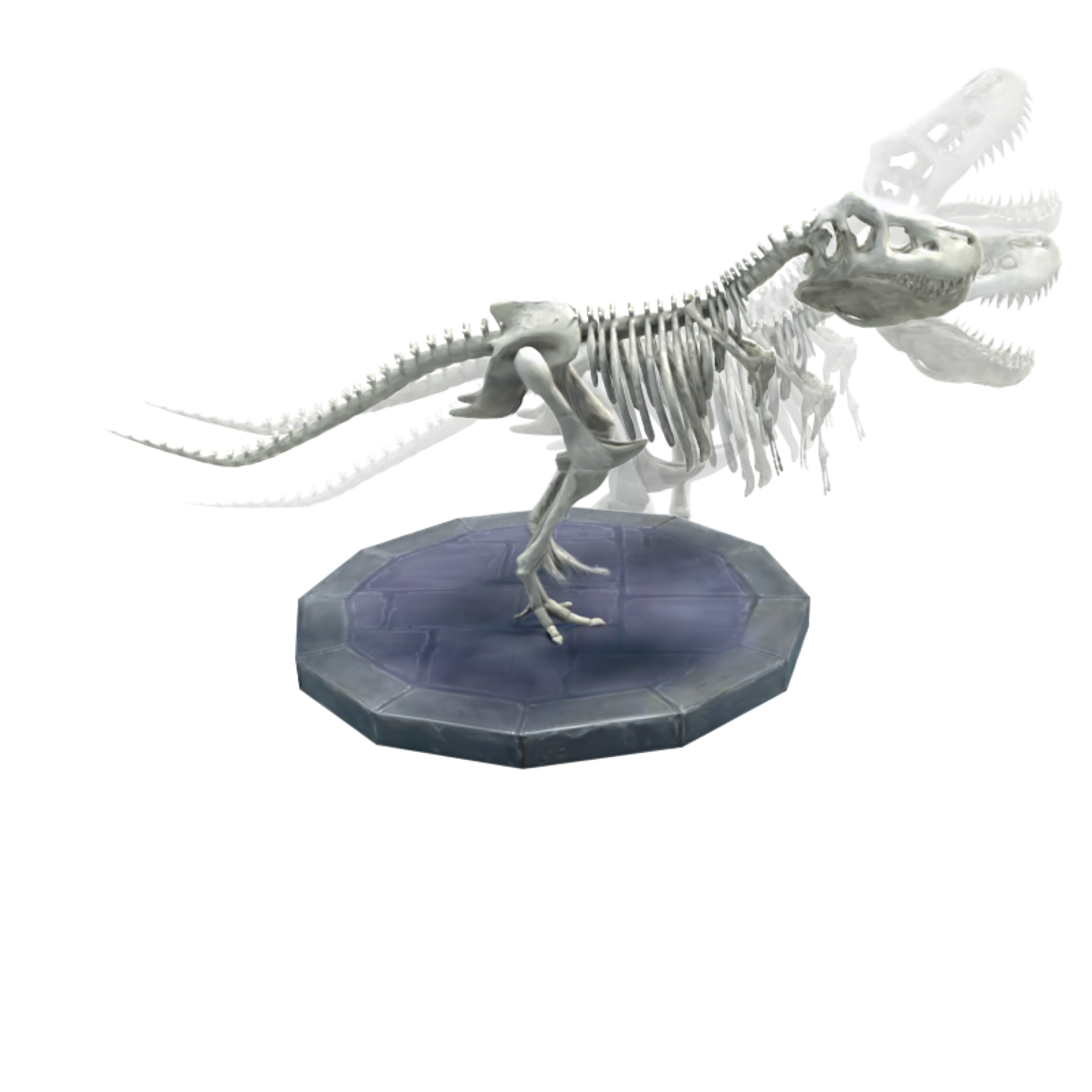} &
    \includegraphics[width=1\linewidth, trim=10 10 10 10,clip]{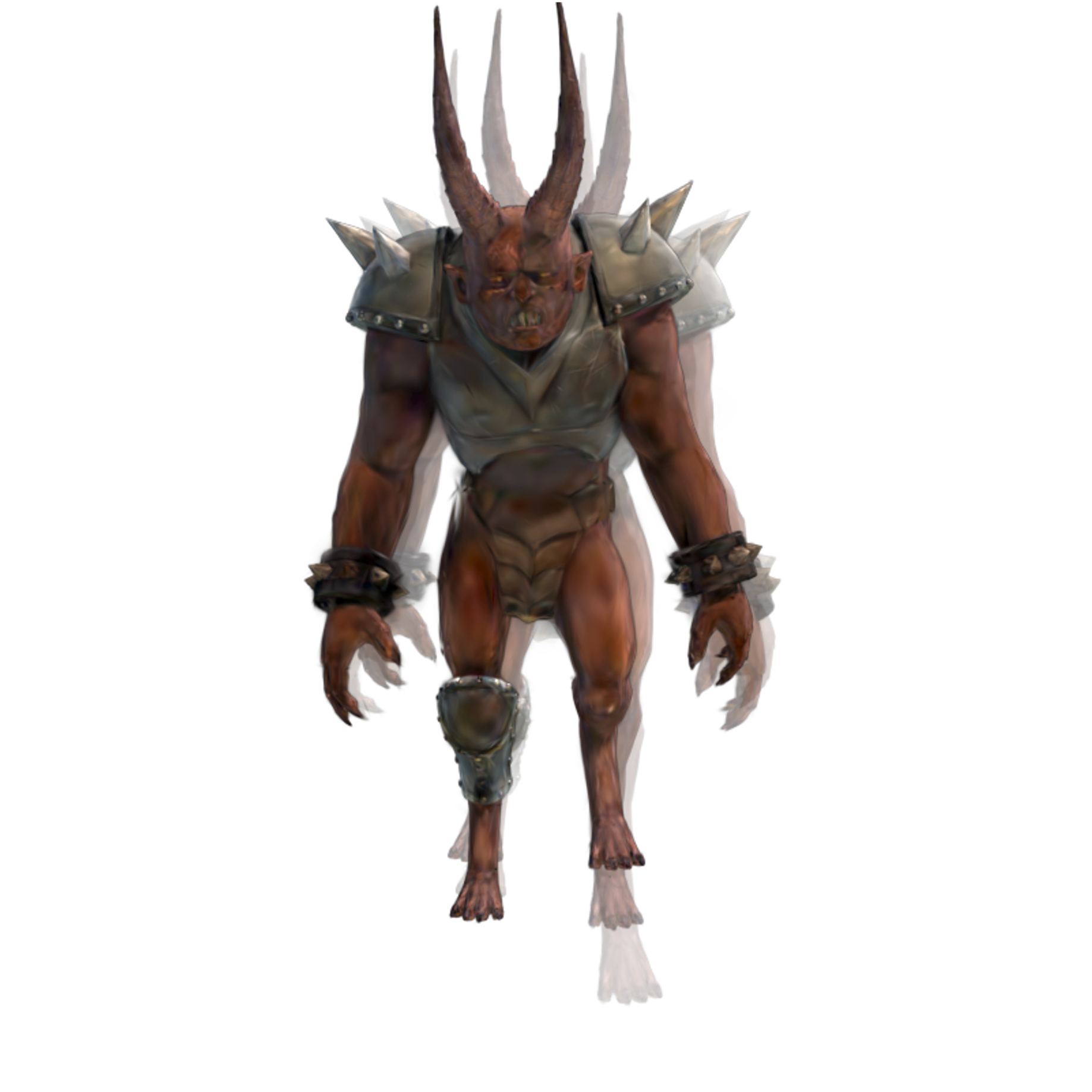}
    \\
    \specialrule{0em}{0pt}{-15pt} \\
    \rotatebox{90}{\quad \small{Edited Motion}} &
    \includegraphics[width=1\linewidth, trim=60 200 490 350,clip]{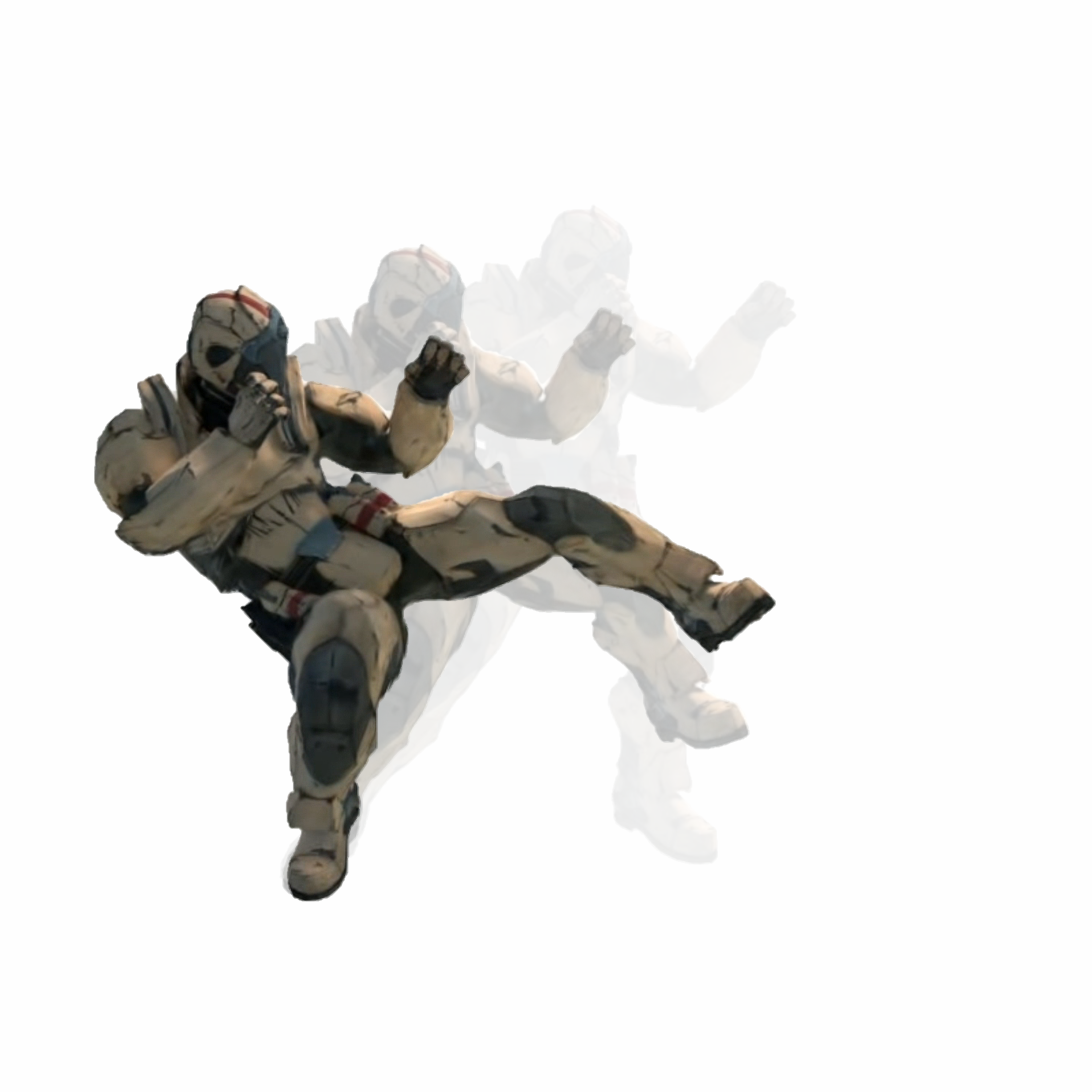} &
    \includegraphics[width=1\linewidth, trim=20 10 50 60,clip]{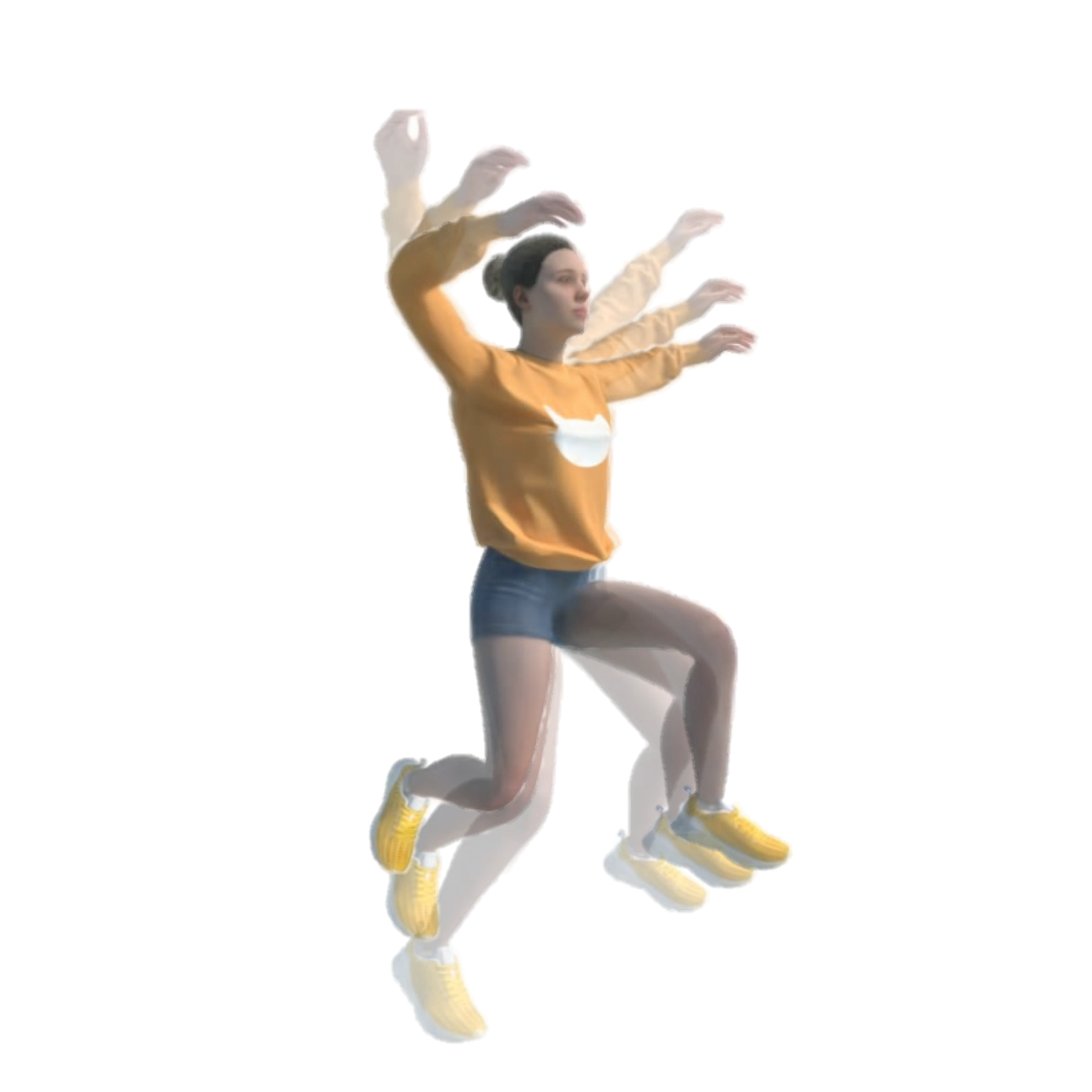}&
    \includegraphics[width=1\linewidth, trim=70 70 60 60,clip]{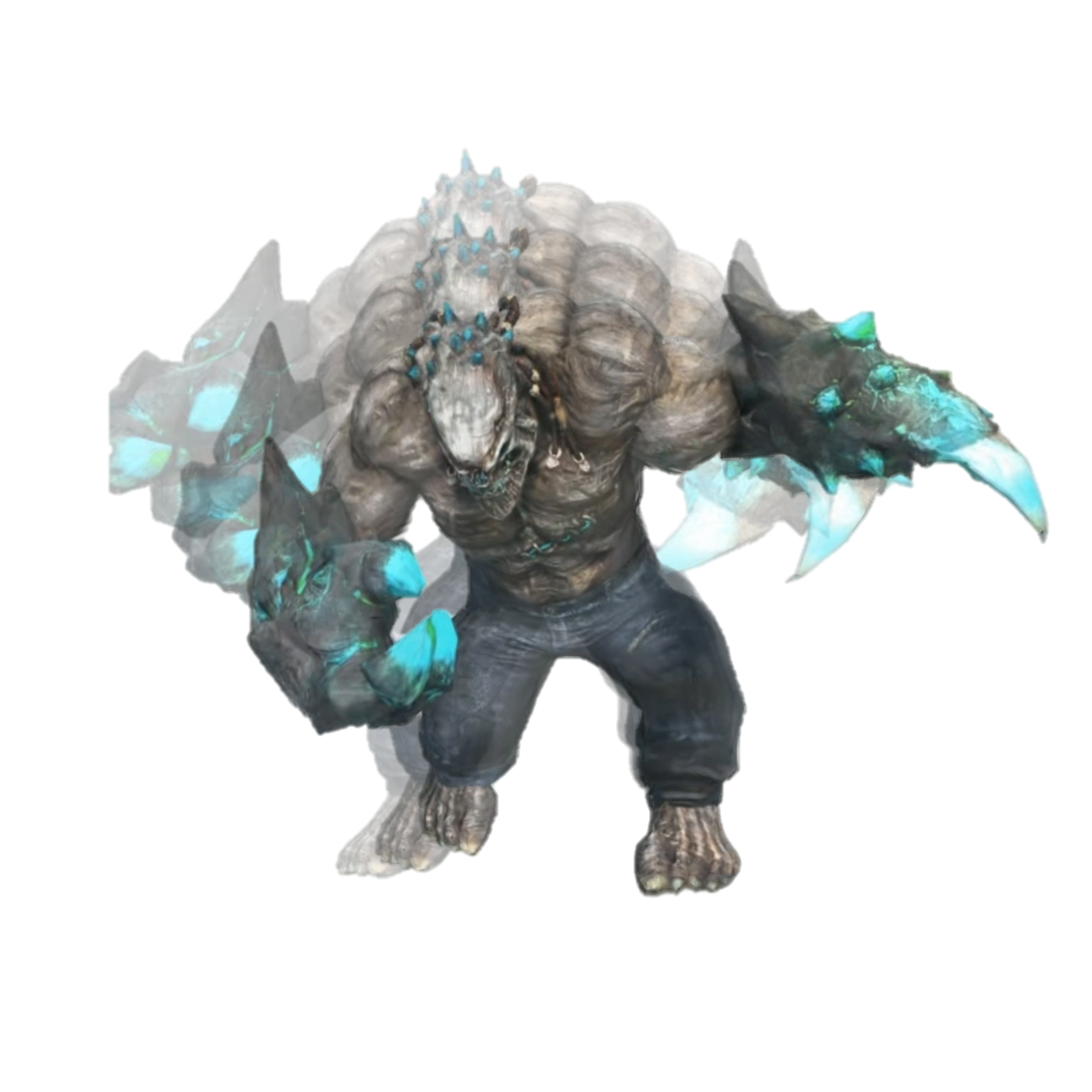}&
    \includegraphics[width=1\linewidth, trim=20 10 50 60,clip]{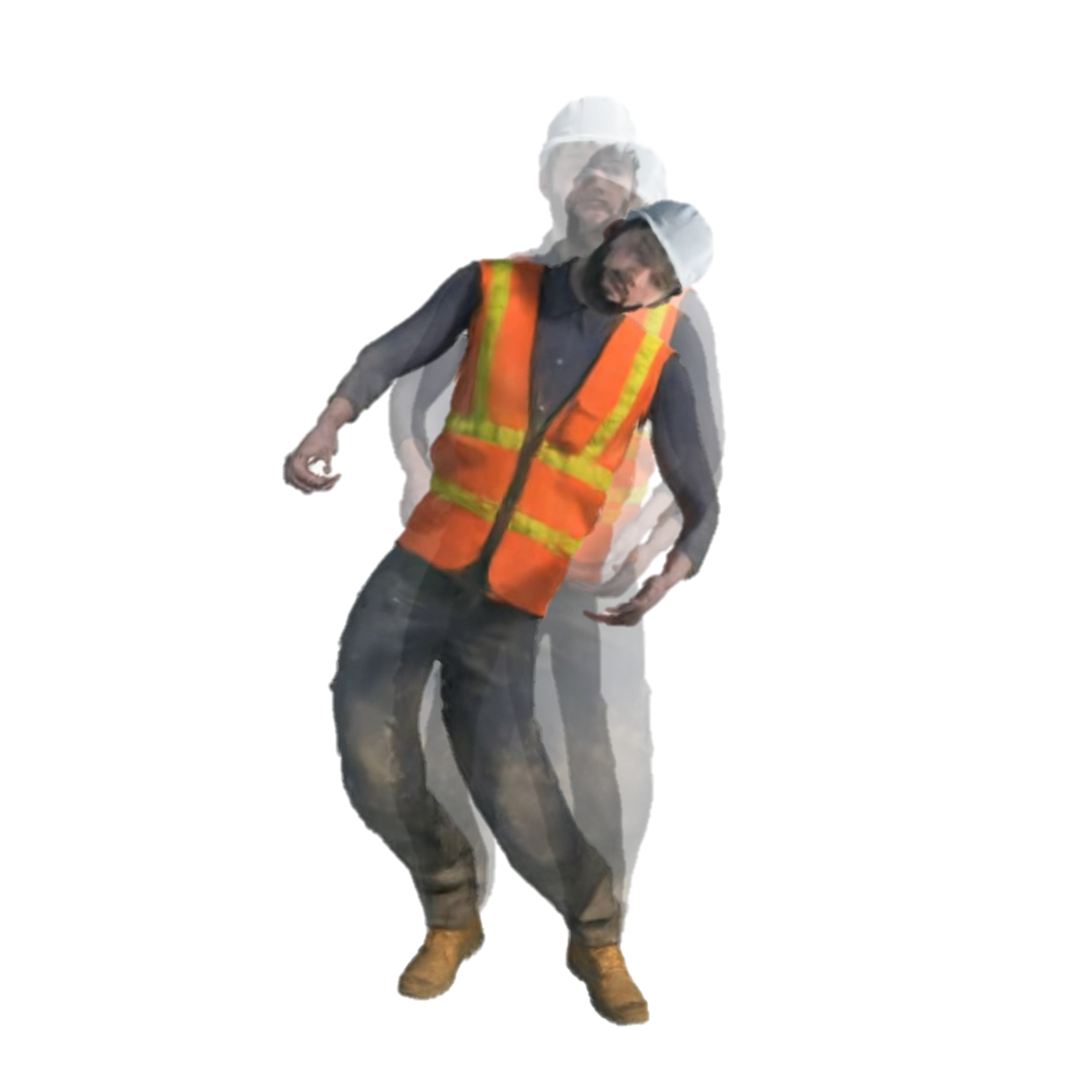}&
    \includegraphics[width=1\linewidth, trim=50 50 0 0,clip]{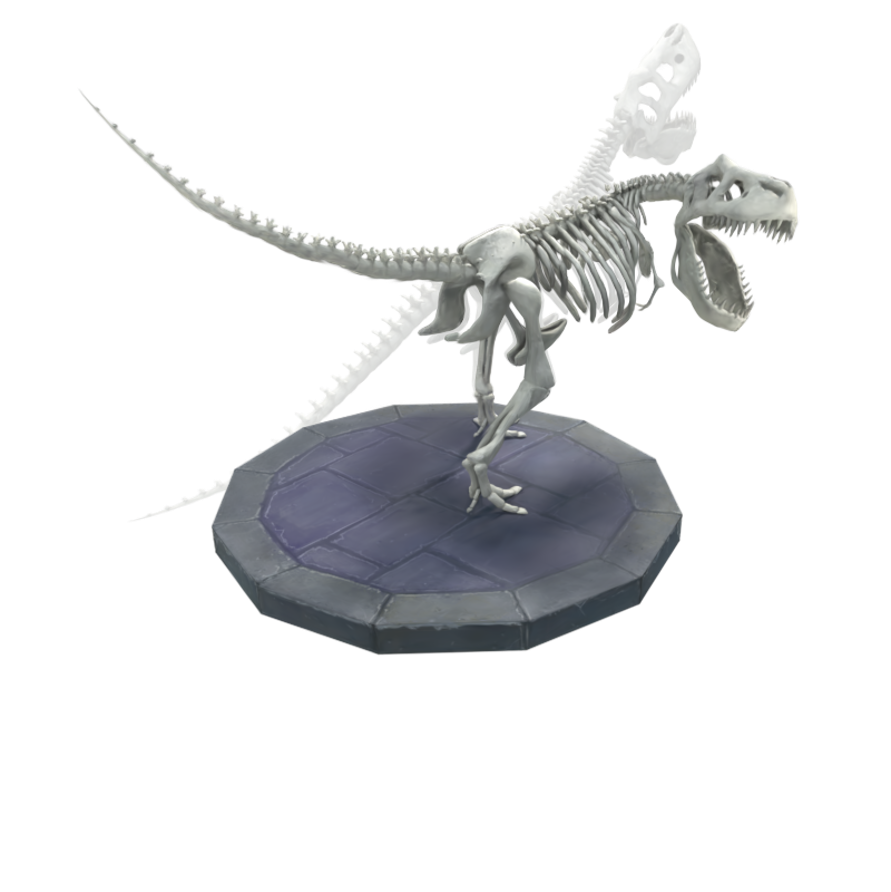}&
    \includegraphics[width=1\linewidth, trim=90 70 60 80,clip]{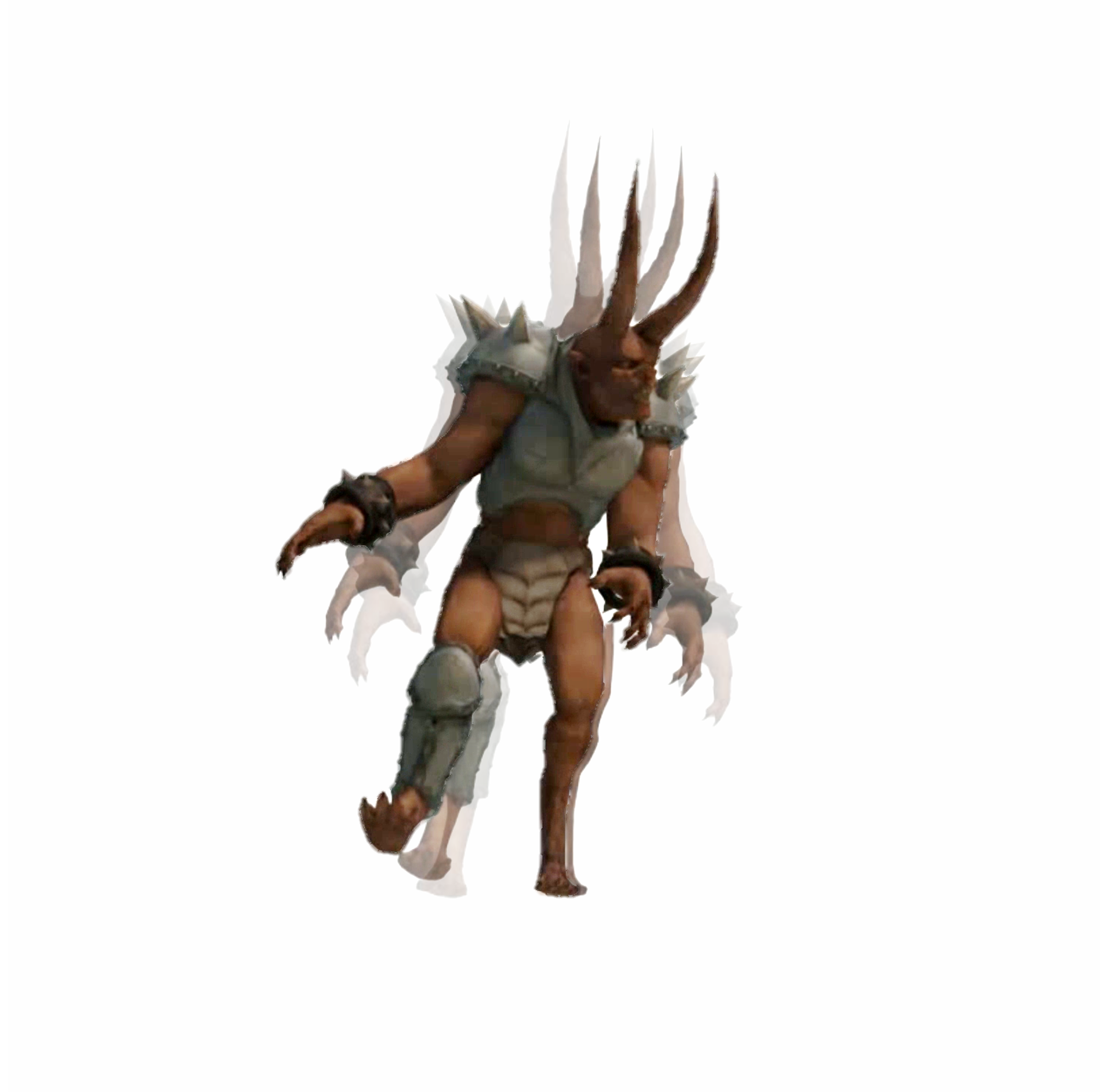}

    \end{tabular}
    \end{spacing}
    \vspace{-9mm}
    \caption{ We visualize the reconstructed motion sequence from the dynamic scene (top) and edited motion sequence (bottom). Our approach generalizes well for motion out of the training set benefitting from the locally rigid motion space modeled by control points.}
    \vspace{-6mm}
    \label{fig:compare_editing}
\end{figure*}

\subsection{Qualitative Comparison}
We also conduct qualitative comparisons to illustrate the advantages of our method over SOTA methods. The comparisons on D-NeRF datasets are shown in Fig.~\ref{fig:compare_sota_synthetic}, where zoom-in images show the details of synthesized images. Our approach produces results closest to the ground truths and has the best visual quality.
Note that, in the Lego scene, the motion in the test set does not align with that in the training set, as indicated in the bottom row of the figure. The same observation can also be seen in \cite{yang2023deformable3dgs}. The qualitative comparisons conducted on the NeRF-DS dataset are also demonstrated in Fig.~\ref{fig:nerf-ds}. It is clear that our method is capable of producing high-fidelity novel views, even in the absence of a specialized design for specular surfaces.

\begin{table}[t]
  \centering
  \caption{We quantitatively evaluate the effect of control points and ARAP loss on D-NeRF~\citep{pumarola2021d} datasets.}
  \label{tab:ablation}
  \vspace{-2mm}
  \setlength\tabcolsep{0pt}
    \begin{tabular*}{\linewidth}{@{\extracolsep{\fill}} lSSSSSSSSSSSSSSS}
        \toprule[1pt]{\footnotesize{Methods}} & {\footnotesize{PSNR($\uparrow$)}} & {\footnotesize{SSIM($\uparrow$)}} & {\footnotesize{LPIPS($\downarrow$)}} \\
          \midrule
        \footnotesize{w/o Control Points}
        & \footnotesize{38.512} & \footnotesize{0.9922} & \footnotesize{0.0162}      \\
        \footnotesize{w/o ARAP loss}
        & \footnotesize{42.617} & \footnotesize{0.9963} & \footnotesize{0.0067}        \\
        \footnotesize{Full}
        & \footnotesize{\textbf{43.307}} & \footnotesize{\textbf{0.9976}} & \footnotesize{\textbf{0.0063}}     \\

        \bottomrule[1pt]
    
      \end{tabular*}
  \vspace{-6mm}
\end{table}

\subsection{Ablation study}

\begin{figure}[tbp]
	\centering
	\includegraphics[width=1\linewidth]{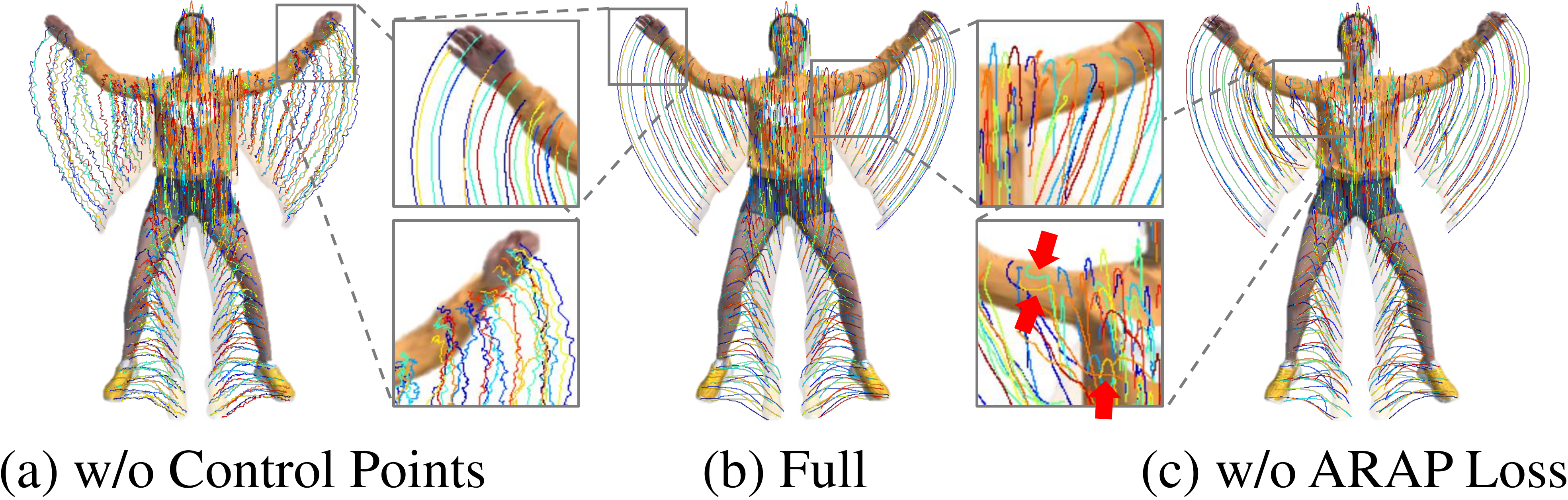}
	\vspace{-6mm}
	\caption{We visualize the rendering results and Gaussian trajectories of (a) the baseline method without control points, (b) our full method, and (c) our method without ARAP loss.}
         \vspace{-5mm}
\label{fig:ablation1}
\end{figure}

\noindent\textbf{Control Points.} Our motion representation driven by control points constructs a compact and sparse motion space, effectively mitigating overfitting on the training set. We quantitatively compare the novel view synthesis quality of our method with the baseline that does not utilize control points on both D-NeRF~\citep{pumarola2021d} and NeRF-DS~\citep{yan2023nerf} datasets, as presented in Tab.~\ref{tab:recon} and Tab.~\ref{tab:nerf_ds_table}. To intuitively elucidate the effects of control points, we compare the results and visualize the trajectories of Gaussians driven either with or without control points in Fig.~\ref{fig:ablation1} (a) and (b). Clearly, directly predicting the motion of each Gaussian with an MLP leads to noise in Gaussian trajectories. 
While the baseline theoretically is more flexible in representing diverse motions, it tends to falter and descend into local minima during optimization, hindering it from achieving the global optimum.

\noindent\textbf{ARAP Loss.} Despite the control-point-driven motion representation providing effective regularization to Gaussian motions, there can be occasional breaches in rigidity. As evidenced in Fig.~\ref{fig:ablation1} (c), even though Gaussians achieve relatively smooth trajectories, some Gaussians on the arm move towards the girl's torso instead of moving alongside the ascending arm. This issue arises due to the lack of constraints on the inter-relation of control points' motions. By imposing ARAP loss on control points, such phenomena are eliminated, thus facilitating a robust motion reconstruction. Tab.~\ref{tab:ablation} illustrates without ARAP loss, the performance of dynamic view synthesis on D-NeRF~\citep{pumarola2021d} slightly decreases.

\subsection{Motion Editing}
Our method facilitates scene motion editing via the manipulation of control nodes, due to the explicit motion representation using control points. The learned correlation and weights between Gaussians and control points enable excellent generalization, even on motion beyond the training sequence. The reconstructed and edited motion sequences are demonstrated in Fig.~\ref{fig:compare_editing}.

\section{Conclusion and Future Works}
We present a method driving 3D Gaussians using control points and a deformation MLP, learnable from dynamic scenes. Our approach, combining a compact motion representation with adaptive learning strategies and rigid constraints, allows high-quality dynamic scene reconstruction and motion editing. Experiments showed our method outperforms existing approaches in the visual quality of synthesized dynamic novel views. However, limitations exist. The performance is prone to inaccurate camera poses, leading to reconstruction failures on datasets with inaccurate poses such as HyperNeRF~\cite{park2021hypernerf}. 
The current approach also faces limitations in handling common specular effects, resulting in limited improvement on NeRF-DS~\cite{yan2023nerf} datasets. To address this, future work could focus on extending the method by incorporating Spec-Gaussian~\cite{yang2024spec} with a specialized specular design. This enhancement would enable more accurate modeling of highlight and mirror effects.
Furthermore, the presence of blurriness in videos with dynamic objects should be considered. To enhance the robustness of the proposed method, incorporating deblurring techniques~\cite{dai2023hybrid,lee2024deblurring} for novel view synthesis can address this issue effectively.

\section*{Acknowledgement}
\noindent This work has been supported by Hong Kong Research Grant Council - Early Career Scheme (Grant No. 27209621), General Research Fund Scheme (Grant No. 17202422), and RGC Matching Fund Scheme (RMGS). Part of the described research work is conducted in the JC STEM Lab of Robotics for Soft Materials funded by The Hong Kong Jockey Club Charities Trust.

{
    \small
    \bibliographystyle{ieeenat_fullname}
    \bibliography{main}
}


\end{document}